%% file: paper.tex
\documentclass[letterpaper, 10 pt, journal, twoside]{IEEEtran} %
\usepackage[utf8]{inputenc}

\usepackage{mathtools}
\usepackage{graphicx}
\usepackage{booktabs}
\usepackage[font=footnotesize]{subfig}
\usepackage[font=footnotesize]{subcaption}
\usepackage{amsfonts}
\usepackage[ruled,vlined,linesnumbered,boxed]{algorithm2e}
\SetInd{0.1em}{0.5em}
\usepackage{array}
\usepackage{multirow, multicol}
\usepackage{color}
\usepackage[bookmarks=false, hidelinks]{hyperref}
\usepackage{setspace} 
\usepackage{pgfplots}
\usepackage{amssymb}
\usepackage{bm}
\usepackage{comment}
\usepackage{soul}

\newcommand*{\Scale}[2][4]{\scalebox{#1}{$\displaystyle #2$}}%
\usepackage{capt-of}
\usepackage[font={footnotesize}]{caption}
\usepackage{balance}

\newcommand{\algrule}[1][.5pt]{\par\vskip.3\baselineskip\hrule height #1\par\vskip.3\baselineskip}
\makeatother
\SetKwInput{KwParams}{Params}
\SetKwInput{KwData}{Global params.}
\SetKw{return}{return}
\SetKw{Break}{break}

\title{
KRRF: Kinodynamic Rapidly-exploring Random Forest algorithm for multi-goal motion planning
}

\author{Petr Je\v{z}ek, Michal Mina\v{r}\'{\i}k, Vojt\v ech Von\' asek$^{*}$ and Robert P\v{e}ni\v{c}ka
\thanks{Manuscript received: June, 21, 2024; Revised September, 3, 2024; Accepted September, 27, 2024.}
\thanks{This paper was recommended for publication by Editor Lucia Pallottino upon evaluation of the Associate Editor and Reviewers' comments.
This work was supported by (organizations/grants which supported the work.)
This work has been supported by the Czech Science Foundation (GA{\v C}R) under project No. 22-24425S, by the European Union under the project Robotics and advanced industrial production (reg. no. CZ.02.01.01/00/22\_008/0004590), and by CTU grant no SGS23/177/OHK3/3T/13.
Computational resources were provided by the e-INFRA
CZ project (ID:90254), supported by the Ministry of Education, Youth and
Sports of the Czech Republic.
}
\thanks{The authors are with the Multi-robot Systems Group, Faculty of Electrical
Engineering, Czech Technical University in Prague, Czech Republic (\protect\url{http://mrs.felk.cvut.cz/}), $^{*}$vonasvoj@fel.cvut.cz
}
\thanks{Digital Object Identifier (DOI): see top of this page.}
}

\def\xiset{\xi_{\mathrm{set}}}
\def\gammaset{\gamma_{\mathrm{set}}}
\def\ftau{\tau_{\text{final}}}

\def\cost{\mathrm{cost}}

\def\AMAX{A_{\mathrm{max}}}

\def\q{q}

\def\qnear{\q_{\text{near}}}
\def\qrand{\q_{\text{rand}}}
\def\qstart{\q_{\text{start}}}

\def\T{\mathcal{T}}

\def\qfrom{\q_{\text{from}}}
\def\qto{\q_{\text{to}}}
\def\qother{\q_{\text{other}}}

\def\qpop{\q_{\text{pop}}}
\def\qbest{\q_{\text{best}}}
\def\RRTSTAR{RRT$^{\star}$}

\def\T{\mathcal{T}}
\def\U{\mathcal{U}}

\def\qactive{\q_{\text{active}}}
\def\tmax{t_{\mathrm{max}}}

\def\nexp{m} 

\def\dist{\varrho}
\def\C{\mathcal{C}}

\def\dist{\varrho}

\newcommand{\Cfree}[0]{\mathcal{C}_{\text{free}}}

\DeclareMathOperator*{\minimize}{\text{minimize}}

\newcommand{\Review}[1]{{#1}}

\newcommand{\PUBLISHEDIN}{IEEE Robotics and Automation Letters}
\newcommand{\DOI}{10.1109/LRA.2024.3478570} 

\usepackage[placement=top,vshift=-7]{background}
\SetBgScale{1.0}
\SetBgContents{\PUBLISHEDIN. PREPRINT VERSION - DO NOT DISTRIBUTE. \href{https://doi.org/\DOI}{DOI \DOI}}
\SetBgColor{black}
\SetBgAngle{0}
\SetBgOpacity{1.0}

\begin{document}

\maketitle

\begin{abstract}
The problem of kinodynamic multi-goal motion planning is to find a trajectory over multiple target locations with an apriori unknown sequence of visits.
The objective is to minimize the cost of the trajectory planned in a cluttered environment for a robot with a kinodynamic motion model.
This problem has yet to be efficiently solved as it combines two NP-hard problems, the Traveling Salesman Problem~(TSP) and the kinodynamic motion planning problem.
\Review{We propose a novel approximate method called Kinodynamic Rapidly-exploring Random Forest~(KRRF) to find a collision-free multi-goal trajectory that satisfies the motion constraints of the robot.}
KRRF simultaneously grows kinodynamic trees from all targets towards all other targets while using the other trees as a heuristic to boost the growth.
Once the target-to-target trajectories are planned, their cost is used to solve the TSP to find the sequence of targets. 
The final multi-goal trajectory satisfying kinodynamic constraints is planned by guiding the RRT-based planner along the target-to-target trajectories in the TSP sequence.
Compared with existing approaches, KRRF provides shorter target-to-target trajectories and final multi-goal trajectories with $1.1-2$ times lower costs while being computationally faster in most test cases.
The method will be published as an open-source library.
\end{abstract}


\vspace{-1.5em}
\section*{Supplementary Material}
{\small
\vspace{-0.4em}
\noindent \textbf{Video:} \url{https://youtu.be/KLneA8Mkep4}\\
\noindent \textbf{Code:} \url{https://github.com/ctu-mrs/krrf}
\vspace{-0.8em}
}

\section{Introduction}

\IEEEPARstart{T}{he} task of multi-goal motion planning involves finding the shortest collision-free trajectory connecting several targets (goals), assuming the order of their visit is unknown in advance. 
Solving this problem is crucial, e.g., in active 
perception~\cite{ mcmahon2015autonomous}, 
multi-robot exploration~\cite{roperoTERRAPathPlanning2019}, and data 
collection~\cite{mcmahon2021autonomous}.
In particular, the data collection and inspection missions' effectiveness is directly defined by the cost of the multi-goal plan which is to be minimized.
Solving the multi-goal motion planning is challenging as one must solve two NP-hard planning problems simultaneously.
The first \textit{combinatorial part} involves finding the sequence of visits as a solution to the Traveling Salesman Problem (TSP). 
The second \textit{continuous part} uses motion planning to find the collision-free trajectory connecting all the goals. 
Both parts are intertwined since the TSP requires knowledge about mutual reachability and trajectory costs between the targets.

In environments with obstacles, finding collision-free trajectories and solving two-point boundary value problems (BVP) for nonholonomic robots is analytically challenging.
Therefore, motion planning is needed to find target-to-target trajectories and their costs to formulate TSP.
The final multi-goal trajectory must be planned to ensure its feasibility under kinodynamic constraints (it cannot be obtained simply by connecting individual target-to-target trajectories due to discontinuities in the targets).

\begin{figure}[!t]
\centering
{\renewcommand{\tabcolsep}{2.5pt}
\renewcommand{\arraystretch}{0.6}
\scriptsize
\begin{tabular}{cc}
\includegraphics[width=0.23\textwidth]{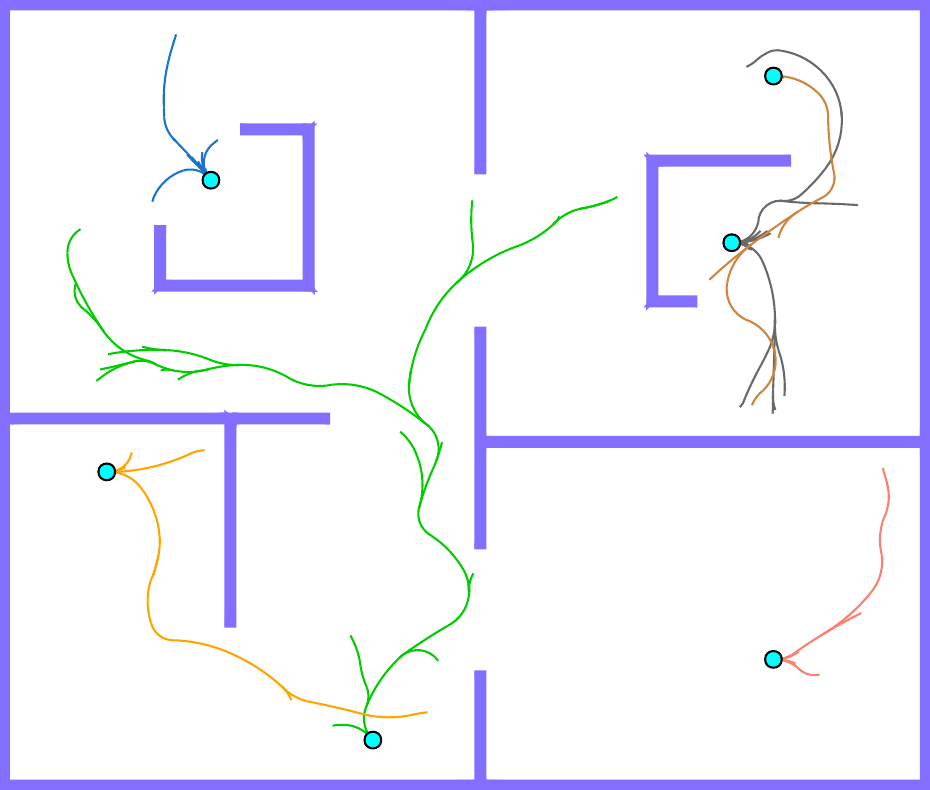} &  \includegraphics[width=0.23\textwidth]{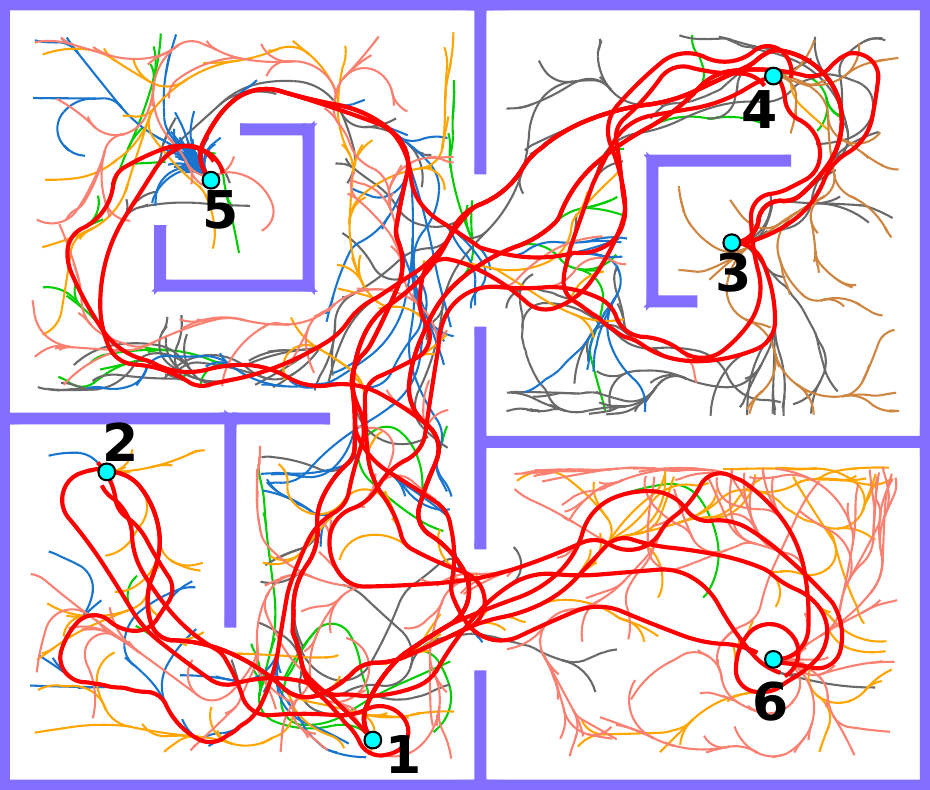}\\
(a) Target-to-target planning & (b) TSP sequence of targets\\ [0.1em]
\includegraphics[width=0.23\textwidth]{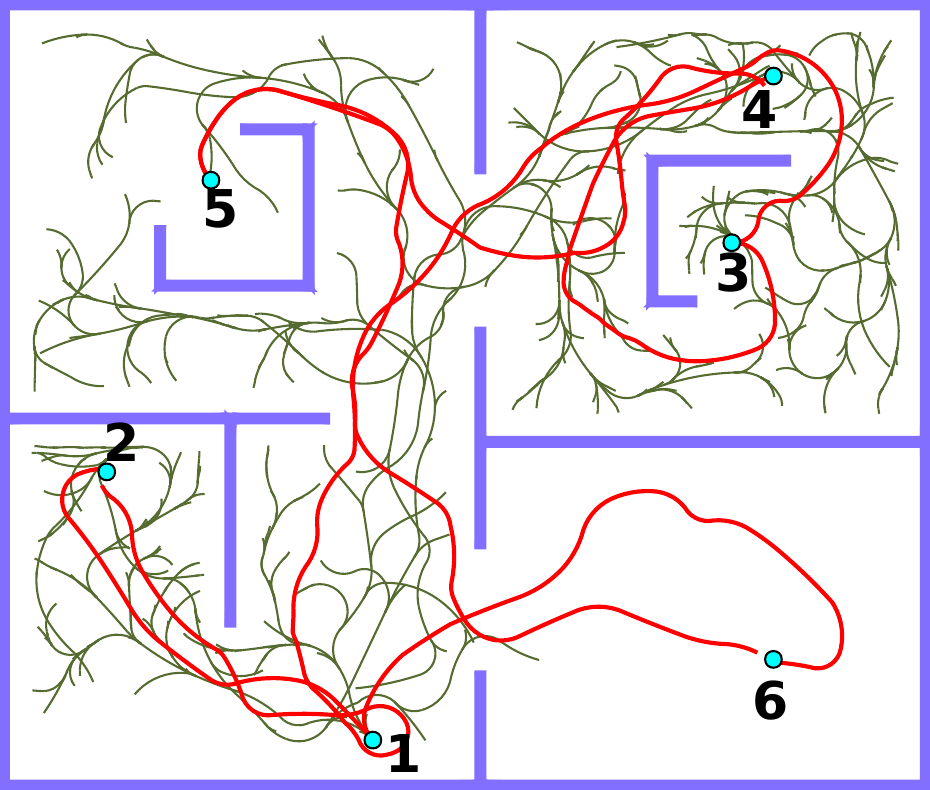} &
\includegraphics[width=0.23\textwidth]{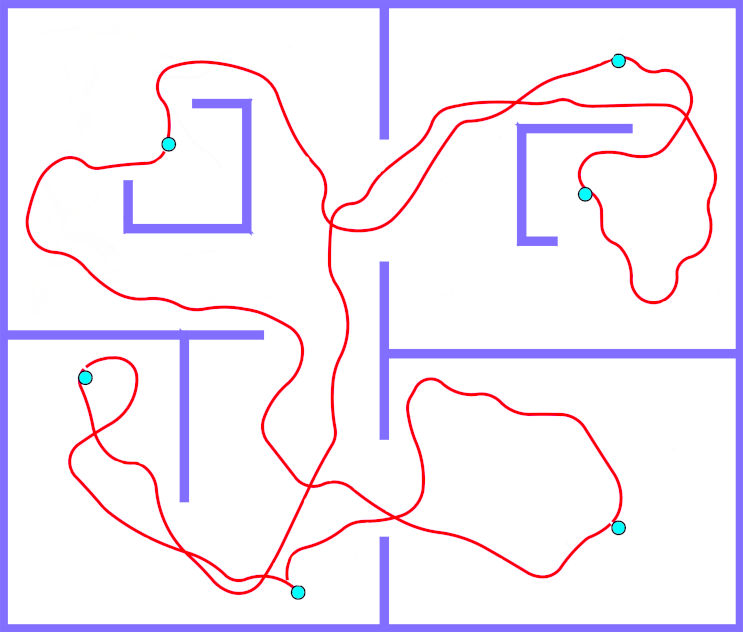}\\
(c) Guided sampling along trajectories & (d) Trajectory through all targets
\end{tabular}
}

\vspace{-0.2em}
\caption{
\Review{
Phases of the KRRF method connecting six targets (circles).
First, target-to-target trajectories are simultaneously planned among all pairs of targets with a  motion planning tree rooted in each target (a).
The minimum-length TSP sequence to visit the targets (marked by numbers) is found based on the lengths of the planned trajectories shown in red (b).
The trajectories connecting the ordered targets (red) guide the RRT-based motion planner (c) to find the final multi-goal trajectory (d).
}
\vspace{-2.4em}
}
\label{fig::intro}
\end{figure}

The multi-goal motion planning problem has not yet been efficiently solved for an arbitrary robot with a kinodynamic motion model in cluttered environments.
A relatively simple approach~\cite{englot2013three}, with up to $n^2$ planning complexity for $n$ targets, is to find the target-to-target trajectories individually, solve TSP, and then connect the previously found trajectories in the TSP order.
\Review{
This, however, is only applicable to robots without a kinodynamic motion model~\cite{englot2013three,janos_multi-goal_2021, Devaurs_TRRT_TSP} or to motion models with a known BVP solution, such as the Dubins vehicle~\cite{janos_randomized_2022}.
However,~\cite{janos_randomized_2022} suffers from 
the discontinuity of the motion model in the targets, leading to additional turns around the targets, thereby increasing the trajectory length.
}

\Review{We propose a novel approximate method called
Kinodynamic Rapidly-exploring Random Forest~(KRRF).
The method grows a motion planning tree from each target simultaneously using 
kinodynamic RRT (Fig.~\ref{fig::intro}a).}
Tree growth is guided by a heuristic that utilizes cost information from one tree to direct the others~\cite{nayak2022bidirectional}. 
After all target-to-target trajectories are found, their cost is used to solve the TSP~\cite{helsgaun2000effective} and to get the optimal order of targets (Fig.~\ref{fig::intro}b).
Finally, the multi-goal trajectory is generated by an RRT-based motion planner, guided by the found target-to-target trajectories (Fig.~\ref{fig::intro}c,d).
\Review{
In comparison to related work, the advantages of KRRF are:
a) we assume a general kinodynamic motion model without a need to solve BVP,
b) tree growth is boosted using a heuristic provided by other trees, which leads to a fast search of 
target-to-target trajectories,
c) the final multi-goal trajectory is feasible, and it satisfies the kinodynamic constraints.
Due to complexity of the multi-goal motion planning, our planner finds an approximate solution, but the experiments show that KRRF provides trajectories with 1.1-2 times smaller costs.
}

\section{Related Work}

\Review{The task of (single-goal) motion planning is to find a feasible trajectory between a start and a goal state~\cite{verasSystematicLiteratureReview2019}.}
Multi-goal motion planning requires, besides constructing the final trajectory, also finding the order of the goals (targets) to be visited so the final trajectory has the lowest cost.
\Review{The most used technique is to formulate the problem of searching the sequence as the Traveling Salesman Problem (TSP)~\cite{spitz2000multiple,saha2003planning,englot2013three,roperoTERRAPathPlanning2019}, or other TSP-related formulations~\cite{khoufiSurveyRecentExtended2019,penicka2019pop}, 
e.g. Close-enough TSP, where a region around each target has to be visited~\cite{yangDoubleloopHybridAlgorithm2018}.}

Solving multi-goal motion planning via TSP requires to define distances between each pair of targets. 
Ideally, the distance should be based on the cost (e.g., length) of a trajectory connecting the targets.
For the Dubins vehicle, the distances can be derived analytically~\cite{penicka2017dubins}, but only in environments without obstacles.
A naive approach to compute the distances in environments with obstacles is to solve motion planning between each pair of targets, which would be time-consuming.

Therefore, several works aim to decrease the computational burden of motion planning.
In Lazy-TSP~\cite{englot2013three}, the initial costs of target connections are based on their Euclidean distances.
An initial tour is computed using TSP, and the connection between the consecutive targets is verified by a time-consuming RRT-based planner.
If no connection is found, TSP is iteratively refined  until a valid sequence and a corresponding trajectory is found.
\Review{Additional speedup can be achieved using learning methods, e.g., to predict collisions of edges~\cite{kew2020batchedmotionplanning}, or to estimate distances between goals using a neural network~\cite{huang2023endtoend}.}
For multi-goal motion planning for fruit inspection,
the method~\cite{kroneman2023afast} avoids computing all target-to-target paths by assuming that each target can be approached from a shell of the tree via an approach path.
This requires computing only $n$ approach paths for $n$ targets.
The distance between two targets is then defined as the sum of the lengths of the approach path plus the length of the path around the shell.




Another way to speed up multi-goal motion planning is to employ faster underlying sampling-based planners.
\Review{These planners are used due to their ability to find motion for various types of robots e.g., mobile or even manipulators~\cite{kejia2023kinodynamic,faroni2023motion}.}
In bidirectional search, two trees are grown until they can be connected~\cite{klemm2015rrtconnect,thakar2022manipulator,devaurs2013enhancing,starek2014bidirectional}.
Connecting the trees requires an analytic solution of BVP~\cite{devaurs2013enhancing,starek2014bidirectional}.
In GBRRT~\cite{nayak2022bidirectional}, forward and backward trees are grown using the forward motion model, which avoids solving the BVP.
The backward tree is used as a heuristic to boost the expansion of the forward tree.
\Review{The work~\cite{faroni2023motion} employs the solution of Multi-Armed Bandit problem to estimate sampling regions for kinodynamic planning of many-DOF robots without BVP solution.}

RRT can be further extended to multiple trees.
\Review{
The configuration space can be explored using multiple (initially disjoint) trees rooted at random configurations~\cite{lai2019balancing}.
If the random sample cannot connect to existing trees, a new tree is created
at the sample. Each random sample is tested for connection to all nearby trees, which ultimately ensures the interconnection of the trees.
The work~\cite{lai2021adaptively} deploys the trees in estimated difficult regions (narrow passages).
Similarly,~\cite{kejia2023kinodynamic} uses multiple trees for motion planning for robotic manipulators and spawns the trees
at random feasible end-effector positions.
}
In our previous work~\cite{janos_multi-goal_2021}, we proposed using multiple trees for multi-goal motion planning. 
The trees are rooted in the targets and expanded using a priority queue, which reflects the distances to other targets.
When two trees approach close enough, they are connected, and multiple connections
between the same trees are permitted.
This approach can be extended to cases where an analytical solution of BVP is known, e.g. for Dubins vehicles~\cite{janos_randomized_2022}.
Multiple trees rooted at the targets are also  used in~\cite{huang20243informable}, but unlike~\cite{janos_multi-goal_2021}, they are grown using \RRTSTAR.
The work~\cite{Devaurs_TRRT_TSP} also uses multiple trees and expands them using a round-robin (in each iteration, a single tree is expanded toward a random configuration).
Unlike~\cite{janos_multi-goal_2021}, only a single connection between each pair of trees is allowed, which results in longer connections (trajectories) between the targets compared to~\cite{janos_multi-goal_2021}.

\Review{
The most relevant works to our solution are works approaching multi-goal motion
planning via the TSP formulation~\cite{kroneman2023afast,thakar2022manipulator,huang20243informable,saeed2021boundary}, but they do not consider kinodynamic constraints.}
Therefore, we propose a new method for multi-goal motion planning assuming kinodynamic constraints. 

\section{Problem Formulation}

\Review{
Finding a minimum-cost collision-free trajectory over multiple target locations, assuming a kinodynamic robot motion model, and with an unknown order of visiting the targets includes two challenging parts:
a) finding feasible trajectories with a minimum cost between all target locations,
and b) the combinatorial optimization problem of the TSP to find the appropriate sequence to visit the targets,  minimizing the overall multi-goal trajectory cost.}
However, due to the kinodynamic robot model, the available target-to-target trajectories can not be simply connected into a final feasible multi-goal trajectory.
Therefore, an additional step to find the final trajectory is needed.

Let $\C$ denote the configuration space and $\Cfree \subseteq \C$ the collision-free region. 
The distance between two configurations $\q,\q' \in \C$ is denoted as $\dist(\q,\q')$ (we assume Euclidean metric).
Robot motion is described using the forward motion model 
$\dot{\q} = f(\q,u), \q \in \C, u \in \U$, where $u$ is a control input from a set of allowed control inputs $\U$.

The robot needs to visit collision-free targets $R = \{r_1, \ldots , r_n\}$ (each is a position in the environment).
We use a circular neighborhood with a radius $R_f$ around each target to be visited, i.e., the target $r_i$ is visited by a configuration $\q$ if $\dist(\q, r_i) \le R_f$.
The sequence to visit the target locations is described by a vector of their indices $\Pi = (\Pi_{1},\ldots,\Pi_{n})$, $1 \le \Pi_{i} \le n$,  $\Pi_{i} \ne \Pi_{j}$ for $i \ne j$.   

The optimal sequence $\Pi$ is found by solving TSP, where 
mutual distances between the targets are computed as a cost of collision-free trajectories connecting the targets.
\Review{A collision-free trajectory between two targets $r_i$ and $r_j$ is  $\tau_{i,j}: [0,1] \to \Cfree$ with $\dist(\tau_{i,j}(0),r_i) \leq R_f$ and $\dist(\tau_{i,j}(1),r_j) \leq R_f$,
let $\cost(\tau_{i,j})$ denote its cost (we use the trajectory length in our experiments as the cost).
}
\Review{Let $\tau_{i,j}^{*}$ denote the optimal  collision-free
trajectory (with the minimum cost) between targets $r_i$ and $r_j$.}
Therefore, the necessary task is to find a set 
$\T^{*}=\{\tau_{i,j}^{*} \; | \; i,j=1,\ldots,n, i \ne j\}$ of collision-free kinodynamic trajectories with minimal costs, which can lead to the optimization problem:
\begin{equation}
\label{opt:tsp_planning}
\begin{aligned}
    \minimize_{\Pi, \T^*} & 
            \sum_{i = 2}^{n} \cost(\tau_{\Pi_{i-1}, \Pi_{i}}^{*}) + \cost(\tau_{\Pi_n, \Pi_{1}}^{*})\\
    \text{subject to: } &
     \Pi_i \in \Pi, 1 \le \Pi_{i} \le n\ , \ \Pi_{i} \ne \Pi_{j} \text{ if } i \ne j\text{, }\\
     & \tau^{*}_{i,j} \in \T^{*} \text{ , }
\Review{      \tau^{*}_{i,j}(t) \in \Cfree, \forall t \in [0,1] \text{,}} \\
     & \dist( \tau^{*}_{i,j}(0), r_i) \leq R_f 
     \text{ , } \dist( \tau^{*}_{i,j}(1), r_j) \leq R_f \text{.}
\end{aligned}
\end{equation}

The solution to the above-formulated problem is a sequence of visits $\Pi$ and
the set of target-to-target trajectories $\T^{*}$. 
The second task of multi-goal motion planning is to find the final trajectory connecting all the targets. 
This final trajectory cannot be made simply by connecting trajectories $\tau_{ij}^{*}$ in a given order as it may violate the kinodynamic constraints.
Moreover, connecting the individual trajectories naively may even yield a discontinuous trajectory since the last configuration of $\tau_{ij}^{*}$ need not be the same as the first configuration of $\tau_{jk}^{*}$.
Therefore, finding the final trajectory requires to solve another optimization problem:
\Review{
\begin{equation}
\label{opt:multigoal}
\begin{aligned}
    \minimize_{\T_{\Pi}} & \,\,\cost(\T_{\Pi}) 
    \!\!=\!\!
    \sum_{i=2}^n 
    \cost(\tau'_{\Pi_{i-1},\Pi_{i}}) \!
    + \!
    \cost(\tau'_{\Pi_{n},\Pi_{1}})  \\
    \text{subject to: } & \tau'_{i,j}   \in \T_{\Pi}, \tau'_{i,j}(t) \in \Cfree, \forall t \in [0,1] \text{,} \\
    & \dist( \tau'_{i,j}(0), r_i ) \leq R_f \text{ , }
     \dist( \tau'_{i,j}(1), r_j ) \leq R_f \text{,} \\
& \tau'_{\Pi_i,\Pi_{i+1}}(1) = \tau'_{\Pi_{i+1},\Pi_{i+2}}(0), 1 \le i < n-1 \text{,} \\
& \tau'_{\Pi_{n-1},\Pi_n}(1) = \tau'_{\Pi_{n},\Pi_1}(0)  \text{.} 
\end{aligned}
\end{equation}
}

The solution to this problem is the final multi-goal
trajectory 
$\T_{\Pi} = (\tau'_{\Pi_1,\Pi_2}, \tau'_{\Pi_2,\Pi_3}, \ldots \tau'_{\Pi_{n},\Pi_1} ) $, which is a continuous feasible sequence of collision-free
trajectories connecting the targets in the order $\Pi$.
While the optimization problems~\eqref{opt:tsp_planning} and~\eqref{opt:multigoal} are written separately, the multi-goal motion planning is a problem that, if to be solved optimally, would require solving them jointly as one influences the other.
\Review{However, the joint solution would be computationally significantly more demanding as it would require, ideally, running the collision-free motion planning through all targets for a large number (similar to~\cite{englot2013three}) of possible sequences of visiting the targets.}
Therefore, as described in the next section, we opted to solve the individual problems separately one after the other.

\section{Proposed Method}

\Review{
The proposed KRRF algorithm for multi-goal kinodynamic motion planning consists of two phases:
}
a) the forest construction phase 
(Alg.~\ref{alg::m3rrf}, lines \ref{alg1:phase1a}--\ref{alg1:phase1b}), and 
b) the trajectory reconstruction phase 
(Alg.~\ref{alg::m3rrf}, lines \ref{alg1:phase2a}--\ref{alg1:phase2b}).
\Review{The construction phase aims to find all target-to-target trajectories, extract their costs to formulate the TSP problem, and solve the TSP problem to obtain the order $\Pi$ of visiting the targets.}
In the reconstruction phase, the final trajectory connecting the targets (given by the order $\Pi$) is computed using the principle of guided sampling.

\subsection{Forest construction}

In the construction phase, all target-to-target trajectories are found, and their lengths are extracted to formulate and solve the related TSP instance.
The configuration space is explored using multiple trees $T_i, i=1,\ldots,n$, each being
rooted at the corresponding target $r_i \in R$.
The trees are pairwise expanded (e.g., $T_i$ and $T_j$, $i<j$).
Each expansion adds a few nodes to the considered trees and is realized using the forward motion model.
Each node holds information about the trajectory length to the root of its tree. 
This information is used to calculate a heuristic that helps to expand nodes towards the roots of the other trees.


\noindent 
{\bf Heuristic}
For two trees $T_i, T_j$, $i<j$, the heuristic $p_{i,j}(\q)$ for $\q \in T_i$ estimates the cost of reaching goal $r_j$ (i.e., the root of tree $T_j$) from the node $\q$,
$p_{i,j}(\q) = \dist(\q,\q_j) + \cost(\q_j,r_j)$, where $\q_j \in T_j$ is a node in the tree $T_j$, and $\cost(\q_j,r_j)$ is the cost (length) of the trajectory in the tree $T_j$ from its node $\q_j$ towards the root $r_j$. 
The node $\q_j \in T_j$ is chosen as the nearest node to $\q$.

The heuristic values $p_{i,j}(\q)$ are stored in the priority queue $Q_{i,j}$.
The priority queue $Q_{i,j}$ is maintained for each pair of trees, $i<j$.
Each element of the queue $Q_{i,j}$ is a pair $( \q, p_{i,j}(\q))$, 
i.e., a node $\q \in T_i$ and its heuristic value. 
The elements in the priority queue are organized
according to the values of $p_{i,j}$ (smaller values have higher priority).

\begin{figure*}[!ht]
\centering
\subfloat[$T_j$ forward expansion]{
\frame{\includegraphics[width=0.23\textwidth]{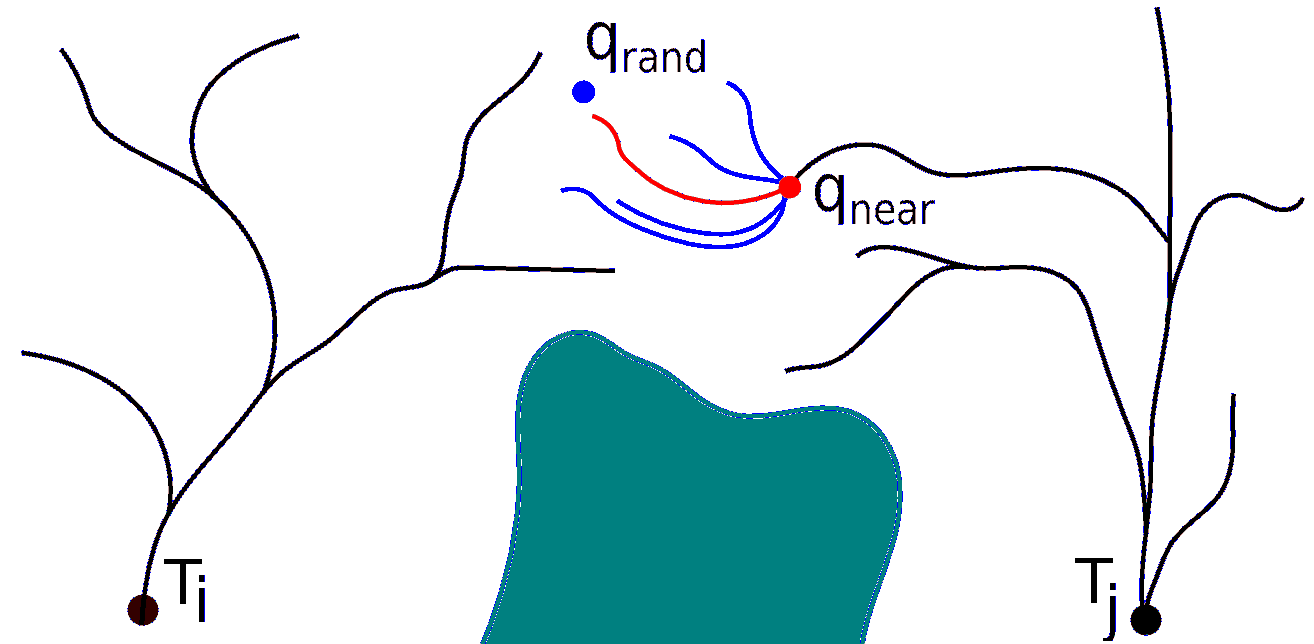}}
}
\subfloat[$T_i$ forward expansion]{
\frame{\includegraphics[width=0.23\textwidth]{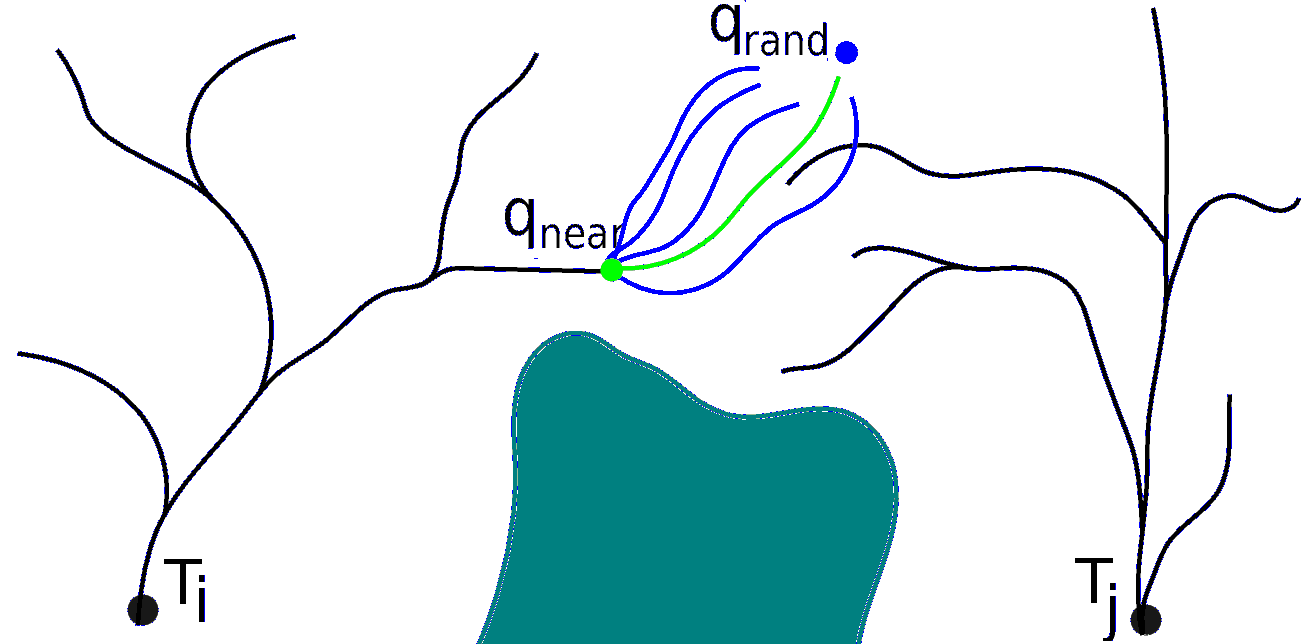}}
}
\subfloat[$T_i$ expansion towards $T_j$]{
\frame{\includegraphics[width=0.23\textwidth]{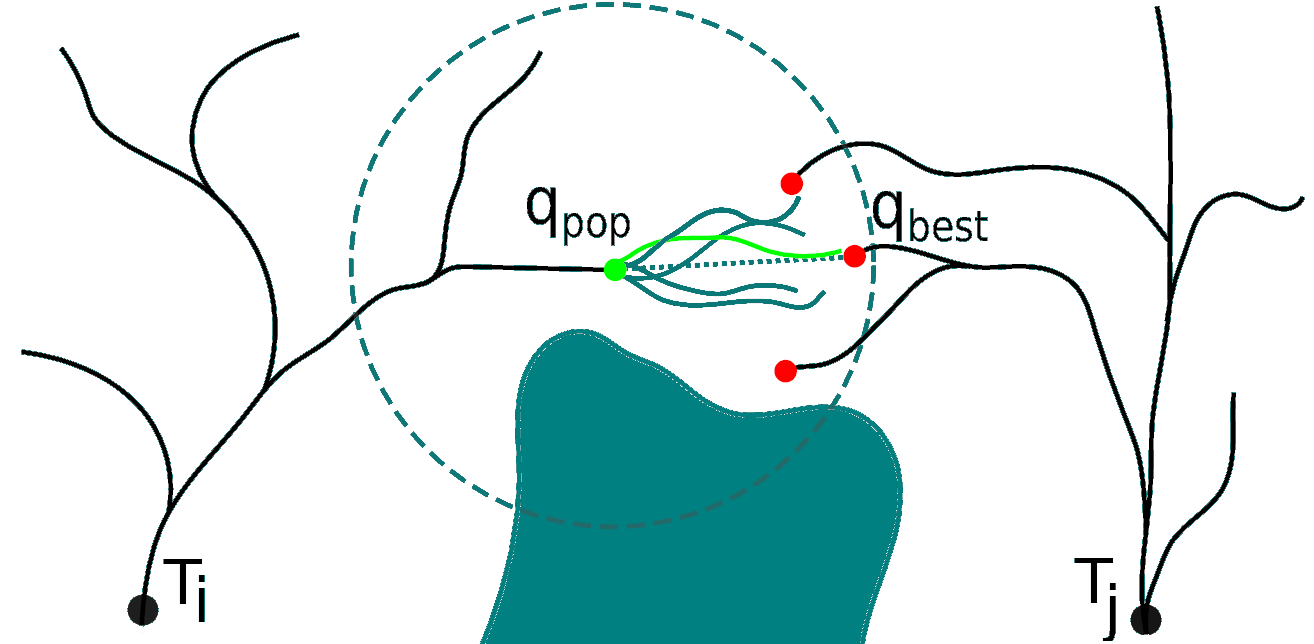}}
}
\subfloat[$T_i$ random expansion]{
\frame{\includegraphics[width=0.23\textwidth]{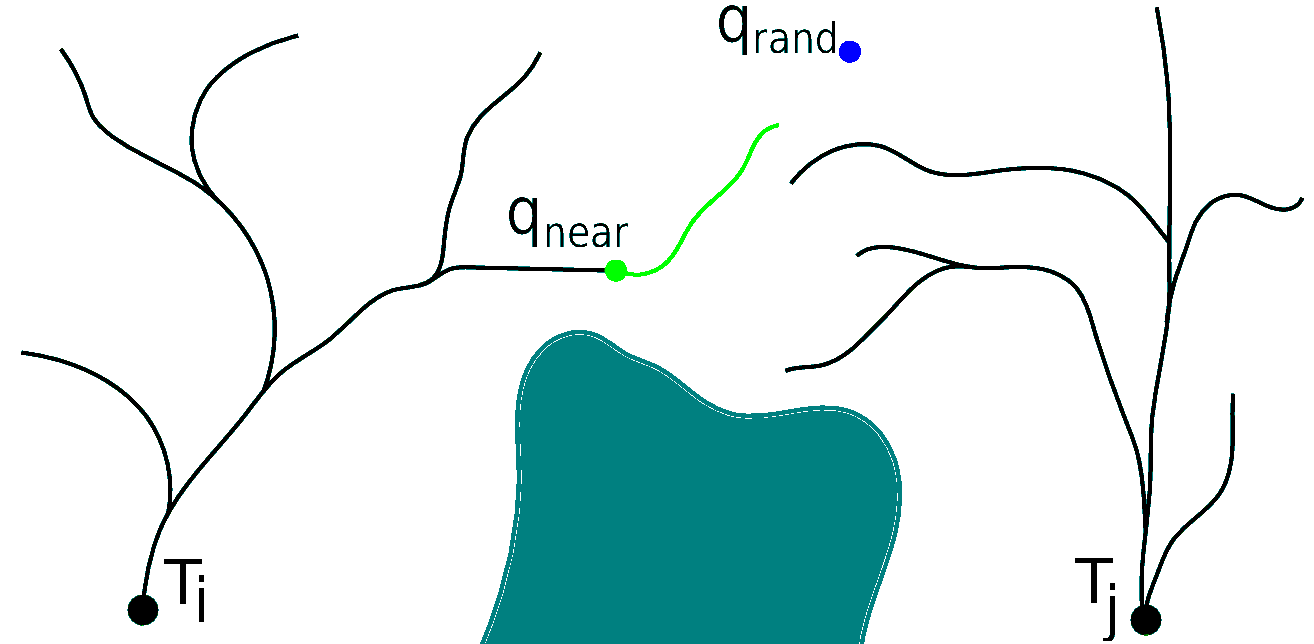}}
}
\vspace{-0.4em}
\caption {
\Review{
Four types of tree expansions. Expansion of tree $T_j$ using Monte Carlo simulation and selecting the nearest configuration to $\qrand$ (red path in (a)). 
Similarly for $T_i$ expansion in (b) green path is selected. 
In (c) an expansion of $T_i$ from node $\qpop$ towards the node $\qbest \in T_j$ in its neighborhood with lowest $\dist(\qpop, \qbest) + \cost(\qbest, r_j)$ (green path).
Random expansion of tree $T_i$ in (d) is just a Monte Carlo simulation with one simulation trial.
}
\vspace{-1.9em}
}
\label{fig::TreeExpansion}
\end{figure*}

\noindent
{\bf Monte Carlo node expansion}
The expansion of a node $\qfrom \in T$ towards a node $\qto \in \C$ is achieved
using Monte Carlo simulation (denoted as 
$\q = T.expansion(\qfrom,\qto,\nexp)$ in the pseudocodes).
The method integrates the kinodynamic model $f(\qfrom, u)$ using a constant
random control input $u \in \U$ applied for a time $t_s$ ($t_s$ is selected
uniformly from $(0, \tmax]$) to obtain a final configuration $\q$.
This process is repeated $\nexp$ times.
The tree is expanded by such a configuration $\q$ that most closely approaches the destination $\qto$ and that is collision-free, i.e., $\q \in \Cfree$ (Fig.~\ref{fig::TreeExpansion}a,b).

\vspace{-0.8em}
\begin{algorithm}
{
\footnotesize
\setstretch{1}
\caption{\label{alg::m3rrf}KRRF}
\KwIn{%
    targets $R = \{r_1, \ldots, r_n\}$; target region $R_f$;
}
\KwOut{%
   trajectory $\ftau$ connecting all targets, or Failure
}
\algrule


\Review{create trees $T_1,  \ldots, T_n$ \tcp*[r]{roots of tree $T_i$ is at $r_i$}} \nllabel{alg1:phase1a}

$Q_{i,j} = \emptyset, \quad \forall i,j = 1,\ldots,n; i < j$ \;
$\tau_{i,j} = \emptyset, \quad \forall i,j = 1,\ldots,n; i \ne j$ \;
\While{ $\exists \tau_{i,j} = \emptyset$ }{ \nllabel{alg1::lineA} 
    $i, j$ = random index pair $i < j \land \tau_{i,j} = \emptyset$\;
    $\tau$ = ExpandTrees$(T_{i}, T_{j},Q_{i,j})$\tcp*[r]{Alg.\ref{alg::expandtrees} \nllabel{alg1::lineB}}
    \If{$\tau \ne \emptyset$}{
        $\tau_{i,j} = \tau$\;
        $\tau_{j,i} = \tau$\;
    } 
}
$D =$ distance matrix from costs of trajectories $\cost(\tau_{i,j})$ \nllabel{alg1:tspmatrix} \;
$\Pi = (\Pi_1, \ldots, \Pi_n) = $ TSP$(D)$ \tcp*[r]{solve TSP} \nllabel{alg1:phase1b} \nllabel{alg1::tsp}
$\qstart = r_{\Pi_1}$ \nllabel{alg1:phase2a}\; 
\Review{$\ftau = \emptyset$; $i = 1$; $a = 0$\;}
\While{\Review{$i \leq n \,\,\mathbf{and}\,\, a < \AMAX $}}
{\nllabel{alg1::lineC}
    \lIf{$i < n$}{
        $\tau = GuidedSampling(\qstart, \tau_{\Pi_i, \Pi_{i+1}}, r_{i+1})$\;
    }\lElse{
        $\tau = GuidedSampling(\qstart, \tau_{\Pi_n, \Pi_{1}}, r_1)$\tcp*[r]{Alg.\ref{alg::guidedsampling}}
    }
    \If{$\tau \neq \emptyset$}{
        $\ftau = \ftau \cup \tau$\;
        $\qstart$ = last state from $\tau$ \;
        $i = i+1$\;
        $a = 0$\;
    }\Else{
        \Review{$a = a + 1$\;}
    }
}\nllabel{alg1::lineD}
\Review{%
\lIf{$a=\AMAX$}{
\Return{Failure}\;
}\lElse{
\Return{$\ftau$} \;
}
}
\nllabel{alg1:phase2b}
}
\end{algorithm}
\vspace{-0.8em}

\noindent
{\bf Growth of the trees}
The trees are expanded incrementally in pairs $T_i$ and $T_j$, $i<j$.
The incremental expansion adds up to one new node to these trees (Alg.~\ref{alg::expandtrees}) and is achieved by expanding selected nodes of $T_i$ and $T_j$ using the Monte Carlo node expansion.
First, the tree $T_j$ is expanded (using Monte Carlo node expansion) 
towards a random sample $\qrand \in \C$ (Alg.~\ref{alg::expandtrees}, line~\ref{alg2::lineD}).
The other tree $T_i$ is expanded from a node that is selected either using Voronoi bias (Alg.~\ref{alg::expandtrees} line~\ref{alg2::lineE} and line~\ref{alg2::lineF}) or using the heuristic, i.e., from a node $\q \in T_i$ that has minimal value $p_{i,j}(\q)$ (Alg.~\ref{alg::expandtrees}, lines~\ref{alg2::lineCH}--\ref{alg2::lineI}).

In the former case, $T_i$ is expanded from a node nearest to a random node $\qrand \in \C$ (Fig.~\ref{fig::TreeExpansion}d).
In the heuristic-based expansion, the most promising node $\qpop \in T_i$ 
is obtained from the queue $Q_{i,j}$.
Then, all nodes in the tree $T_j$ in the radius $h_r$ around $\qpop$ are determined.
From these nodes, a node $\q' \in T_j$ with the minimal value of the heuristic: 
$\dist(\qpop, \q') + \cost(\q',r_j)$ is selected.
The tree $T_i$ is then expanded (using Monte Carlo node expansion)  from $\qpop$ towards $\q'$ (Fig.~\ref{fig::TreeExpansion}c).
This heuristic expansion is based on the GBRRT~\cite{nayak2022bidirectional}
algorithm.
While the original GBRRT solves classical single-goal motion planning, here we use
the principle of GBRRT for multi-goal path planning, and we grow multiple trees instead
of only two (as in~\cite{nayak2022bidirectional}).
The growth of trees $T_i$ and $T_j$, $i<j$ is terminated if the tree $T_i$ approaches
the target $r_j$ close enough (Alg.~\ref{alg::expandtrees}, line~\ref{alg2::lineC}).
Then, the trajectory $\tau_{i,j}$ is extracted and saved for the TSP matrix
computation (Alg.~\ref{alg::m3rrf}, line~\ref{alg1:tspmatrix}).
The trajectory $\tau_{i,j}$ is also used as the trajectory $\tau_{j,i}$ and its length is thus used to estimate the distance
between targets $r_j$ and $r_i$. 

The construction phase is terminated if all target-to-target trajectories are found. 
A target $r_i \in R$ is considered as approached by a tree if the target's distance to any of the tree's nodes is less than the threshold $R_f$ (Alg.~\ref{alg::expandtrees}, line~\ref{alg2::lineC}).

\vspace{-0.8em}
\begin{algorithm}
{\footnotesize
\setstretch{1}
\caption{\label{alg::expandtrees}ExpandTrees}
\KwIn{%
trees $T_{i}$ and $T_{j}$; 
priority queue $Q_{i, j}$;
}
\KwOut{%
trajectory from $r_{i}$ towards $r_{j}$, or $\emptyset$
}
\KwParams{%
$\nexp$: number of Monte Carlo simulations; 
$\gammaset$: exploration parameter; 
$h_r$: radius for updating heuristic;
$R_f$: target neighborhood size; 
}
\algrule
$\qrand^j = $ random from $\C$\;\nllabel{alg2::lineJrand} 
$\qnear^j =$ nearest to $\qrand^j$ in $T_j$\;
$\q = T_j.expansion(\qnear^j, \qrand^j, \nexp)$\tcp*[r]{Fig.~\ref{fig::TreeExpansion}a}\nllabel{alg2::lineD}  
$\qother$ = nearest to $\q$ in $T_i$\;
$p_{i,j}(\qother) = \dist(\qother, \q) + \cost(\q,r_j)$ \;
update or push element $(\qother, p_{i,j}(\qother))$ to $Q_{i,j}$\; \nllabel{alg2::push1}
$\qrand^i = $ random from $\C$\;\nllabel{alg2::lineIrand}
$\qnear^i =$ nearest to $\qrand^i$ in $T_i$\; 
\If{$random(0,1) \leq \gammaset$}{
    \If{$Q_{i, j} = \emptyset$}{ \nllabel{alg2::lineH}
    $\q = T_i.expansion(\qnear^i, \qrand^i, \nexp)$\tcp*[r]{Fig.\ref{fig::TreeExpansion}b}\nllabel{alg2::lineE}
    } \Else{
    $\qpop =$ \text{pop element from} $Q_{i, j}$ \; \nllabel{alg2::lineCH}
    $Q = $ nodes from $T_j$ in radius $h_r$ from $\qpop$\;\nllabel{alg2::heuristicA}
    $\qbest = \q' \in Q$ with minimal $\dist({\qpop,\q'}) + \cost({\q',r_j})$\;\nllabel{alg2::heuristicB}
    $\q = T_i.expansion(\qpop,\qbest, \nexp)$\tcp*[r]{Fig.\ref{fig::TreeExpansion}c} \nllabel{alg2::lineI} 
    }
}\Else{
    $\q = T_i.expansion(\qnear^i, \qrand^i, 1)$\tcp*[r]{Fig.\ref{fig::TreeExpansion}d} \nllabel{alg2::lineF}
}
$\qother$ = nearest to $\q$ in $T_j$\;
$p_{i,j}(\q) = \dist(\q,\qother) + \cost(\qother,r_j)$ \;
update or push element $(\q, p_{i,j}(\q))$ to $Q_{i,j}$\; \nllabel{alg2::push2}
%
\If{$\dist(\q,r_j) \leq R_f$}{ \nllabel{alg2::lineC}
    \Return{trajectory from root of $T_i$ to $q \in T_i$}\tcp*[r]{traj. $\tau_{i,j}$}
}
\Return{$\emptyset$}
}\end{algorithm}
\vspace{-0.8em}

\noindent
{\bf Priority queue update}
The priority queue $Q_{i,j}$ contains pairs $(\q, p_{i,j}(\q))$.
The update of the queue either adds a new element into the queue (if no element with the configuration $\q$ is in the queue), or finds the element with the configuration $\q$, and updates its heuristic (if the new heuristic value is smaller than the actual value).
After tree $T_j$ is expanded, the queue $Q_{i,j}$ is updated (Alg.~\ref{alg::expandtrees}, line~\ref{alg2::push1}).
Conversely, after tree $T_i$ is expanded, the queue is updated (Alg.~\ref{alg::expandtrees}, line~\ref{alg2::push2}).

\subsection{Multi-goal trajectory reconstruction}

When all target-to-target trajectories $\tau_{i,j}$ are computed, 
their lengths are used in the TSP cost matrix, and the TSP
instance is solved using~\cite{helsgaun2000effective} to find 
the sequence $\Pi$ minimizing the sum of trajectory lengths.
The final multi-goal trajectory $\ftau$ connecting all targets $R$ in the order $\Pi$. 
is computed using guided-based motion planning~\cite{dennyDynamicRegionbiasedRapidlyexploring2020,vonasek2009rrt}.

In guided motion planning, the configuration space is sampled along a `guide' --- in this case, along the trajectories $\tau_{\Pi_i, \Pi_{i+1}}$ that were found in the first phase of the algorithm.

The guiding trajectory is a sequence of configurations.
The guided-based planning builds a configuration tree, and the random samples
are generated around an active waypoint $\qactive$ on the trajectory
with the probability $\xi_{set}$ (guiding bias) (Alg.~\ref{alg::guidedsampling}, line~\ref{alg3::gbias}).
Otherwise, the random sample is generated from $\C$.
When the tree approaches the active waypoint $\qactive$ within $R_f$ distance, the next
waypoint from the guiding trajectory is selected until the last point of the 
guiding trajectory is reached.
The result of the guided-based planning is a trajectory connecting
two consecutive targets $\Pi_i$ and $\Pi_{i+1}$. 
This trajectory is added to the final trajectory $\ftau$ and the planning continues until all targets in the order $\Pi$ are connected (Alg.~\ref{alg::m3rrf} lines~\ref{alg1::lineC}--\ref{alg1::lineD}).

\vspace{-0.5em}
\begin{algorithm}
{
\footnotesize
\setstretch{1}
\caption{\label{alg::guidedsampling}GuidedSampling}
\KwIn{%
    starting node $\qstart$; 
    guiding trajectory $\tau$; 
    target $r$;
}
\KwOut{%
    trajectory from $\qstart$ to $r$ \;
}
\KwParams{%
$\nexp$: number of Monte Carlo simulations; 
$\xiset$: goal-bias for guided sampling; 
$R_f$: target neighborhood size; 
$k$: maximum number of guiding iterations
}
\algrule
initialize tree $T$ with $\qstart$ \;
$\qactive = $ first node in $\tau$\;
\For{$i = 1, \dots, k$}{
    \If{$random(0,1) < \xiset$}{
    $\qrand =$ random node around $\qactive$: $\dist(\qrand, \qactive) \leq R_f$ \; \nllabel{alg3::gbias}
    }\Else{
    $\qrand =$ random from $\C$\;
    }
    $\qnear =$ nearest to $\qrand$ in $T$\;
    $\q = T.expansion(\qnear, \qrand, \nexp)$ \;
    \lIf{$\dist(\q,r) \leq R_f$}{
        \Return trajectory from $\qstart$ to $\q$ in $T$\;
    }
    \lIf{$\dist(\q,\qactive) \leq R_f$}{
        $\qactive = $ next state in $\tau$ \;
    }
}
\Return{$\emptyset$}\;
}
\end{algorithm}
\vspace{-1.0em}

\subsection{Discussion}
\label{sec:Discussion in method}

The advantage of exploring the space using multiple trees is two-fold:
a) 
Unlike bidirectional search, where the trees need to be connected by solving a (time-consuming) two-point boundary problem~\cite{starek2014bidirectional,devaurs2013enhancing}, the proposed method does not connect the trees (and therefore, no BVP solution is needed). 
Instead, the trees grow using the forward motion model until they approach all the targets.
b) Information about configuration space contained in one tree (i.e., trajectory lengths of the tree's nodes towards its root) is used as a heuristic that helps to  select nodes of the other trees for expansion. 
By using the heuristic, the trees prefer to expand from promising nodes, which speeds up reaching the targets.
\Review{KRRF relies on sampling-based search of the configuration space, which is known to be able to deliver plans for wide range or robots including mobile robots or robotic manipulators~\cite{kejia2023kinodynamic}.
While we further present only results for kinodynamic models in 2D worlds, the proposed planner is generally applicable also to vehicles operating in 3D and for many-DOF vehicles (thanks to fact, that the trees are built by RRT principle which can search even high-dimensional configuration space).

Since the trees are grown using non-optimal RRT-based planners, the estimation of costs (i.e., $\mathrm{cost}(\tau_{i,j})$), is not optimal. 
The final trajectory is also constructed using non-optimal sampling-based planner. 
Therefore, the proposed approach is an approximate method for multi-goal motion planning, but it ensures that the final trajectories are feasible for the kinodynamic robots. 
It may happen that the final multi-goal trajectory cannot be constructed in the given order of targets.
In such a case, the planner attempts to perform up to $\AMAX$ connection between consecutive targets (Alg.~\ref{alg::m3rrf}, line~\ref{alg1::lineC}), before reporting failure. 
}

Several parameters influence the performance of KRRF.
The parameter $\nexp$ determines how many Monte Carlo simulations are used to expand a single node. 
Higher values of $\nexp$ lead to a slower computation and should be used for motion models with many control inputs and in densely cluttered environments.
We found $\nexp = 10$ satisfactory (higher values did not result in shorter trajectories).
The length of the Monte Carlo simulation is determined by the parameter $\tmax$: longer duration may lead to faster tree expansion but may be counterproductive in highly cluttered environments. 
Moreover, $\tmax$ should be chosen considering the motion model. 
We recommend setting $\tmax$ to such a value that the simulated trajectories are not longer than $R_f$ (so the node expansion does not `overshoot' the desired goals).

The radius $h_r$ determines how close two trees need to be to update the heuristic (and elements in the priority queues). 
Low values of $h_r$ lead to a low likelihood of updating the heuristic, which slows down the growth of the trees.
On the contrary, a larger radius $h_r$ leads to more frequent updates of the heuristic. 
We recommend setting this parameter to the value $2R_f$.
The parameter $\gammaset$ (Alg.~\ref{alg::expandtrees}) defines the
probability of using the heuristic expansion.
Generally, a higher $\gammaset$ favors exploitation (growing trees more likely along other trees) over exploration (growing trees uniformly in the configuration space).
According to~\cite{nayak2022bidirectional}, $\gammaset > 0.5$, which was also used in our experiments.

The guiding process (Alg.~\ref{alg::guidedsampling}) is influenced by the guiding bias $\xiset$: low values of $\xiset$ prefer uniform sampling, while values approaching one lead to a strict sampling of the guiding trajectory.
We recommend setting $\xiset = 0.9$ as the guiding trajectory is already feasible.
\Review{The guiding is repeated $\AMAX$ times (we used $\AMAX=5n$ in the experiments), and it needs to generate up to $k$ random samples (Alg.~\ref{alg::guidedsampling}).}
We recommend setting this parameter to at least the number of samples used to build the largest tree.

\begin{figure*}[!ht]
\newcommand{\height}{90pt}
\centering
{\renewcommand{\tabcolsep}{6.5pt}
\renewcommand{\arraystretch}{0.6}
\footnotesize
\begin{tabular}{cccc}
\includegraphics[height=\height]{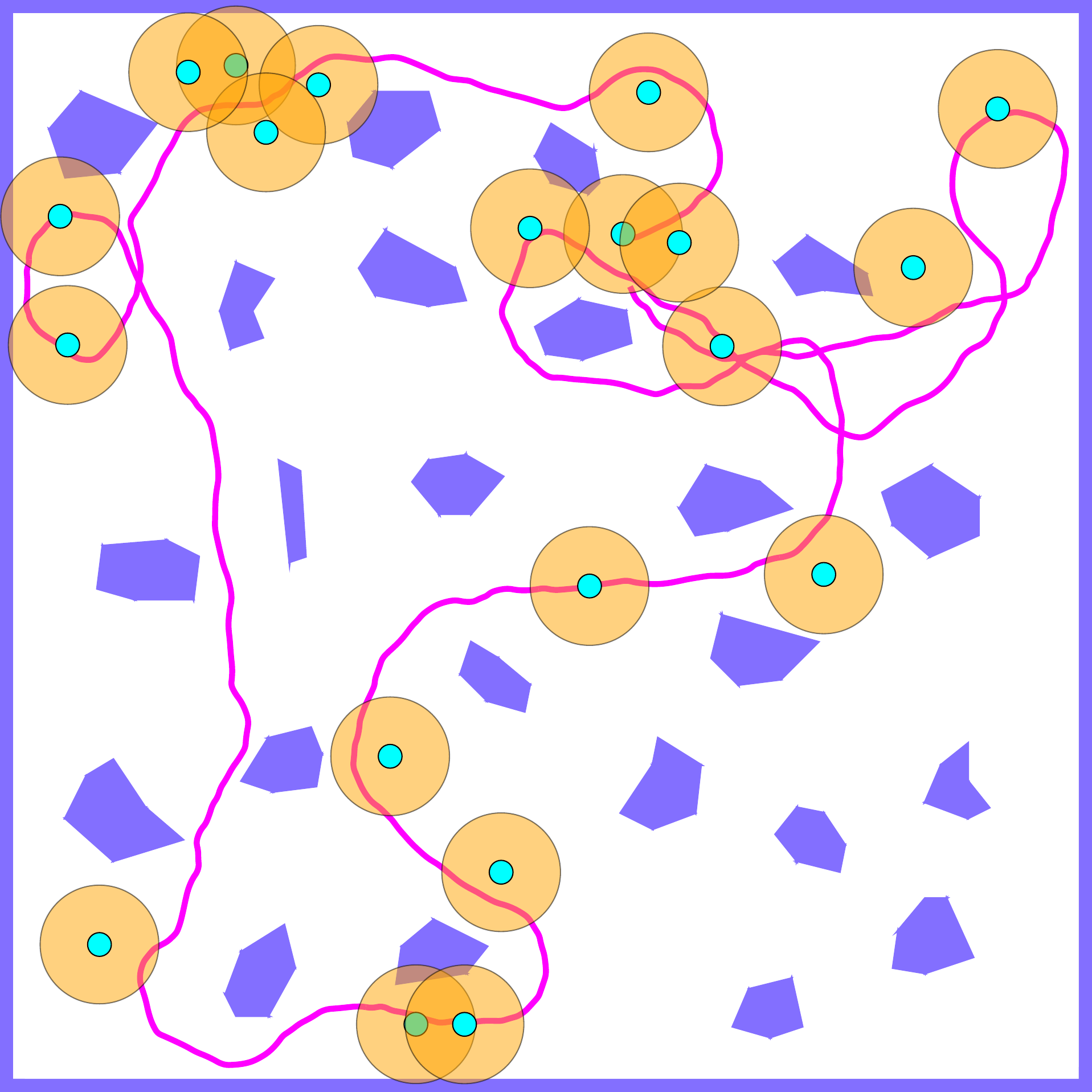} &
\includegraphics[height=\height]{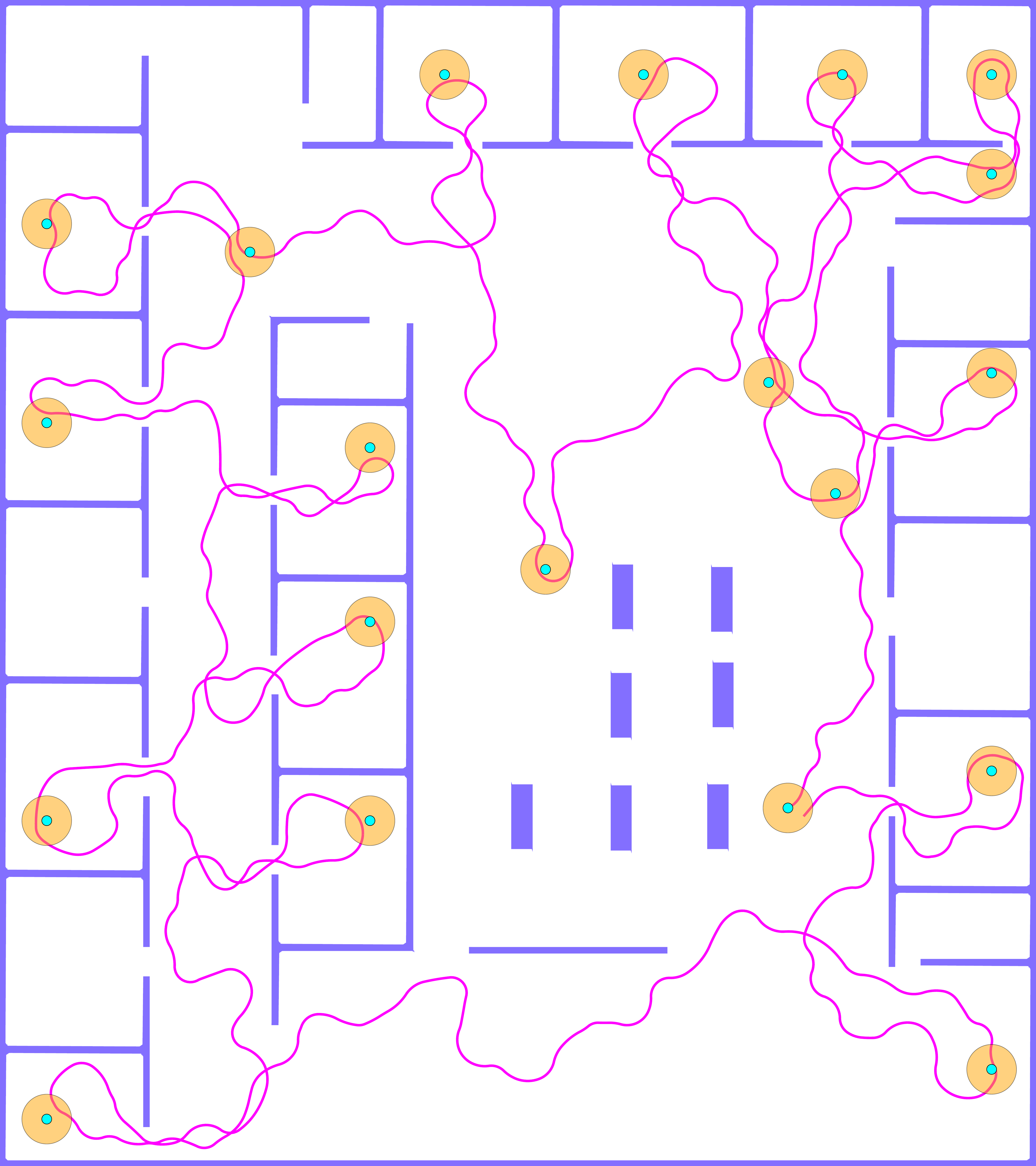} &
\includegraphics[height=\height]{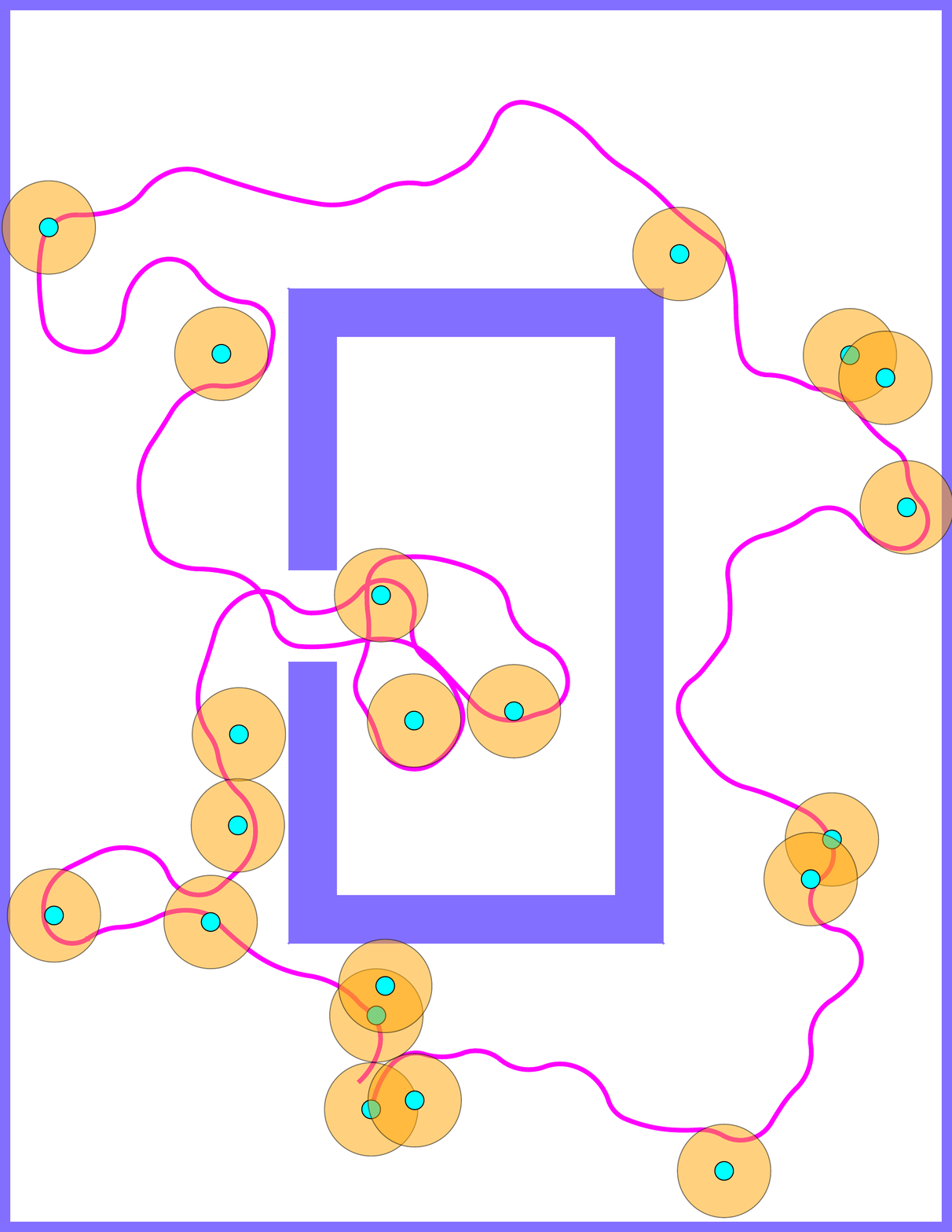} &
\includegraphics[height=\height]{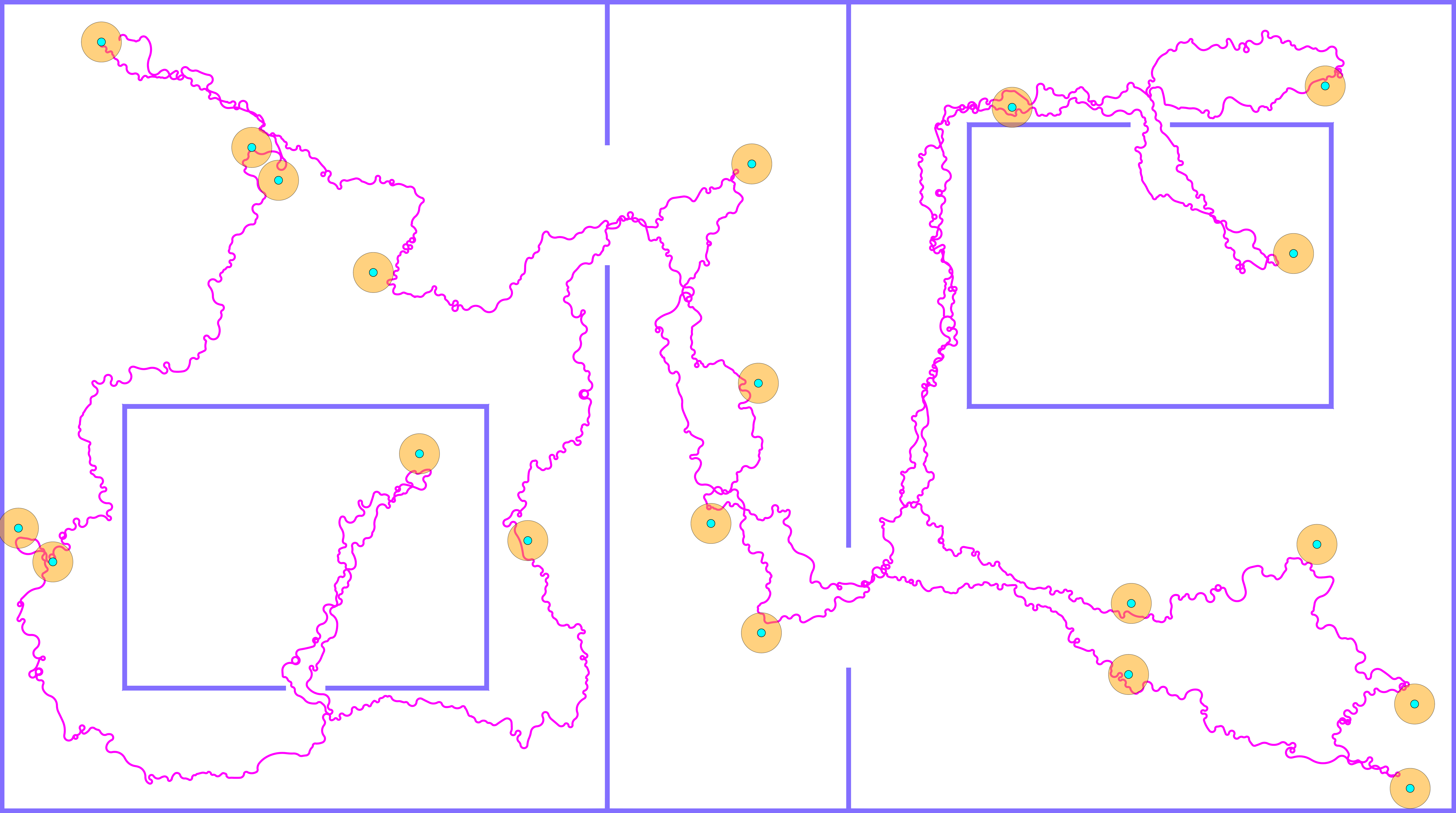}
\\
(a) Potholes 20, KRRF, Diff & 
(b) Floor 20, KRRF, Car &
(c) BT 20, KRRF, Car &
(d) DoubleBT 20, KRRF, Bike \\[0.2em]
\includegraphics[height=\height]{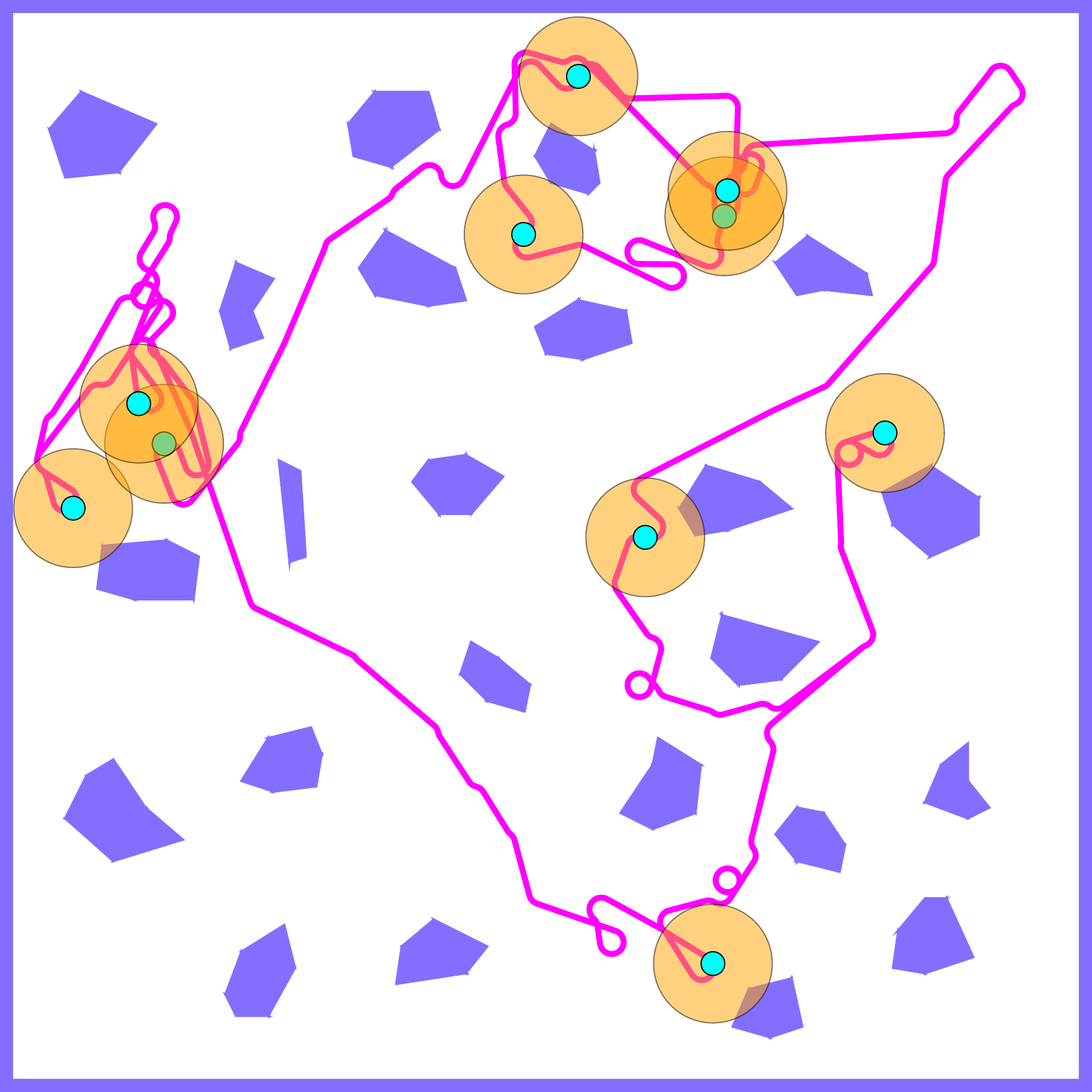} &
\includegraphics[height=\height]{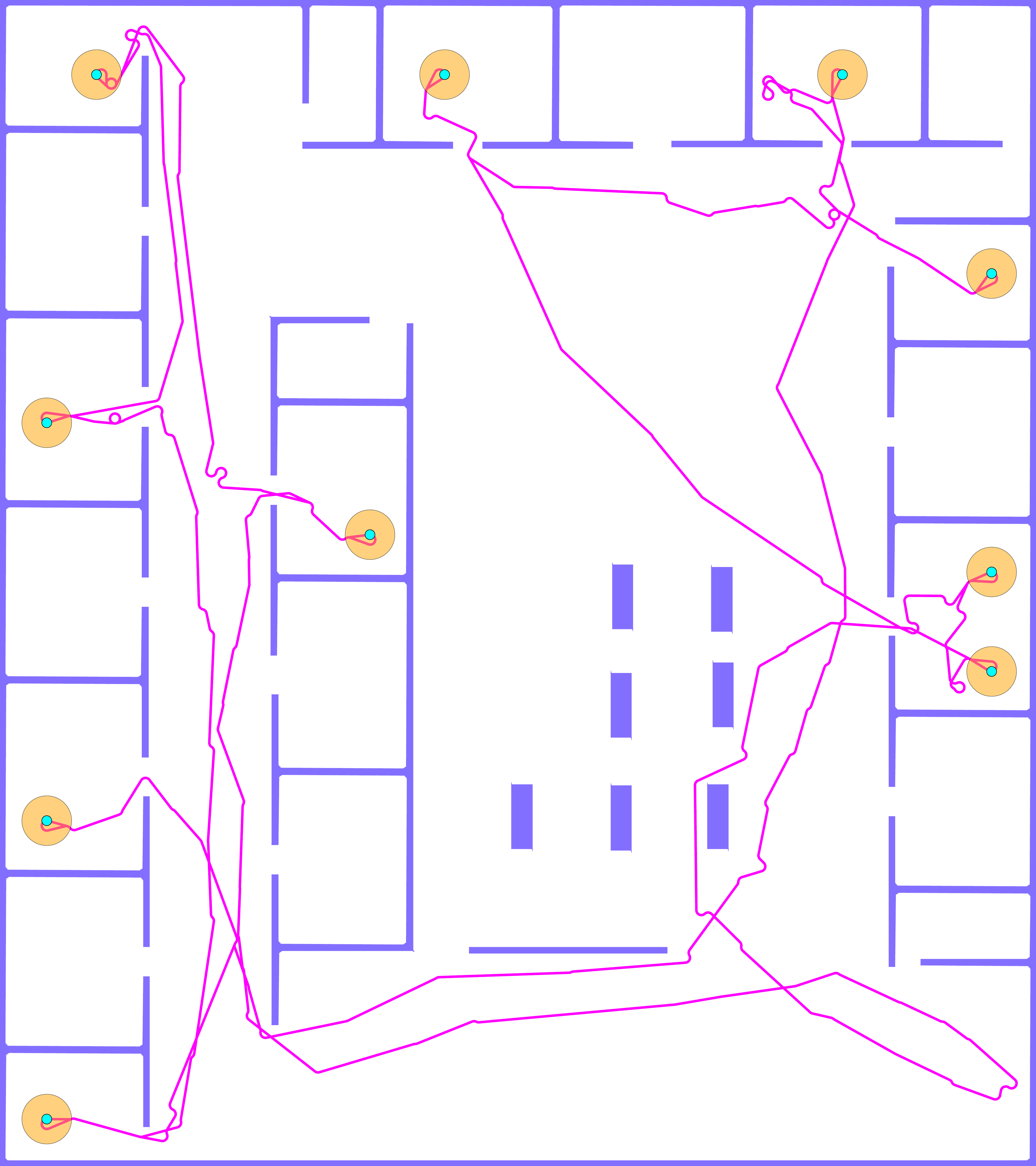} & 
\includegraphics[height=\height]{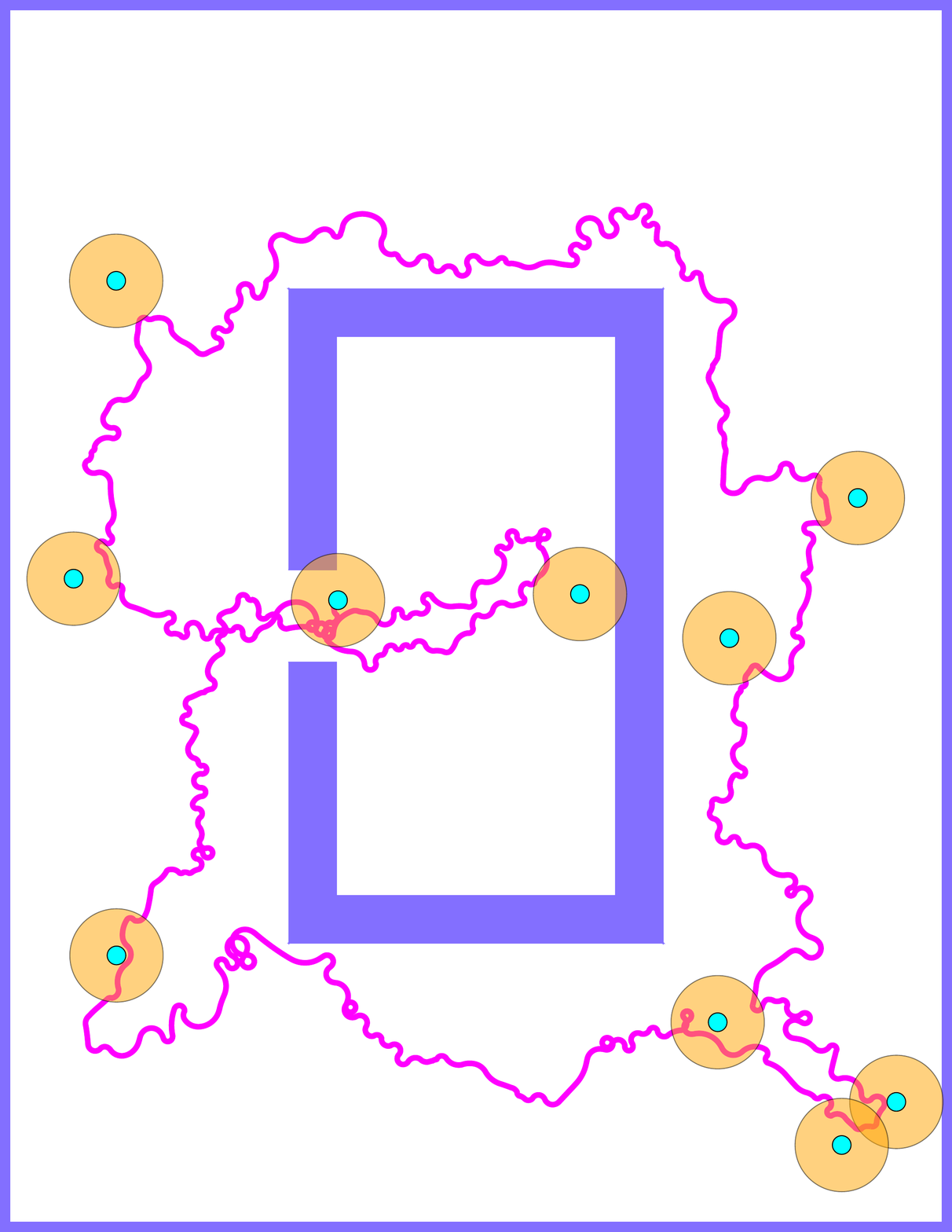} &
\includegraphics[height=\height]{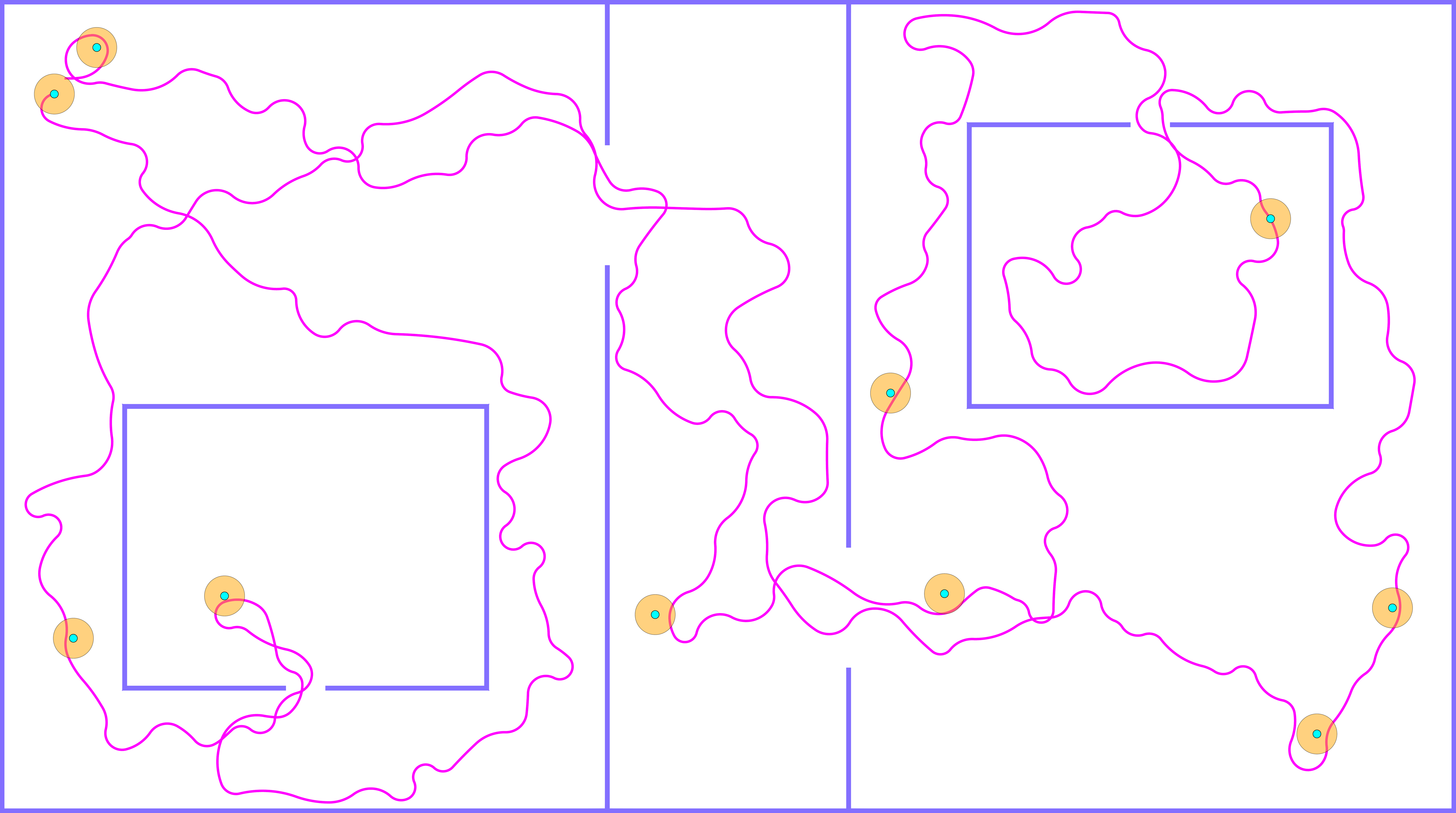}
\\
(e) Potholes 10, SFF*, Dubins &
(f) Floor 10, SFF*, Dubins & 
(g) BT 10, LazyTSP, Bike & 
(h) DoubleBT 10, LazyTSP, Car
\end{tabular}
}
\vspace{-0.4em}
\caption{
Final multi-goal trajectories on Potholes, Floor, Bugtrap (BT), and Double Bugtrap (DoubleBT) maps for tested algorithms and robots. The orange circle represents the region $R_f = 50$ map units. 
}
\label{fig::map_intro}
\vspace{-1.8em}
\end{figure*}

\section{Results}

In this section, the experimental results of KRRF algorithm are presented.
First, the dynamic models of robots used in the experiments are described. 
The models are then further used for a comparison of KRRF with two state-of-the-art algorithms  SFF*~\cite{janos_randomized_2022} and LazyTSP~\cite{englot2013three}.
All methods are implemented in C++, and the experiments are run on an 8-core i7-10700 2.90GHz processor with 32 GB of RAM.
The proposed KRRF was run with the settings: $R_f = h_r = 50$ map units,
$\tmax = 1.5$~s, $\nexp = 10$, $\gammaset = 0.7$, $\xiset = 0.9$, $k=25 \times 10^3$.

\vspace{-1em}
\subsection{Used kinodynamic models}

Three dynamic models of robots were used in the experiments.
The first model is a \textit{Car-like} model controlled by two inputs $u_s$ and $u_{\phi}$, setting the speed and steering angle, respectively.
The dynamic model is represented by the equations
$\dot{x} =  u_s\cos{\theta} $,
$\dot{y} =  u_s\sin{\theta}$, and 
$\dot{\theta} = \frac{u_s}{L}\tan{u_{\phi}}$,
where the states $x$, $y$, and $\theta$ are the position and rotation of the robot in a 2D environment and $L$ is the distance between the front and rear axles.
For the experiments, we limit the inputs to $u_s \in [0,50]$ and $u_{\phi} \in [-\frac{\pi}{4}, \frac{\pi}{4}]$.
The distance $L$ was set to $30$ and robot of size $20 \times 20$ map units.
When the speed is set to a constant, the model is called the \textit{Dubins vehicle} which is the robot motion model used in 
SFF*~\cite{janos_randomized_2022}.

The second model is a \textit{Differential drive} (Diff-drive) model with inputs $u_l$ and $u_r$, controlling the angular velocity of the left and right wheel, respectively.
The dynamical model is:
$\dot{x} = 0.5r(u_l + u_r)\cos{\theta}$,
$\dot{y} = 0.5r(u_l + u_r)\sin{\theta}$, and
$\dot{\theta} = (r/L)(u_r - u_l)$,
where $x$, $y$, and $\theta$ are the position and rotation in 2D, $L$ is the distance between the left and right wheel, and $r$ is the wheel radius.
For experiments, we chose $u_l$, $u_r \in [0,2]$, $L = 20$, $r = 20$, and the size of the robot was $20\times20$ map units.

The third model is a \textit{Bicycle-like} (Bike-like) model~\cite{ge_numerically_2020}.
Inputs to the model are $a$ and $\delta$, the longitudinal acceleration and steering angle.
The model is expressed by the following discretized equation:

\noindent
\begin{equation}\label{eq:bike-like}
\begin{aligned} 
x_{k+1} &=  x_k + T_s(u_k\cos{\theta_k} - v_k\sin(\theta_k)) \\ 
y_{k+1} &=  y_k + T_s(v_k\cos{\theta_k} + u_k\sin(\theta_k)) \\
\theta_{k+1} &= \theta_k + T_s\omega_k \\
u_{k+1} &= u_k + T_sa_k  \\
v_{k+1} &= \Scale[0.8]{\frac{mu_kv_k+T_s(l_fk_f-l_rk_r)\omega_k - T_sk_f\delta_ku_k-T_smu_k^2\omega_k}{mu_k - T_s(k_f+k_r)}} \\
\omega_{k+1} &= \Scale[0.95]{\frac{I_zu_k\omega_k+T_s(l_fk_f-l_rk_r)v_k - T_sl_fk_f\delta_ku_k}{I_zu_k - T_s(l_f^2k_f+l_r^2k_r)}}
\end{aligned}
\end{equation}
where $x$, $y$, and $\theta$ are the position and rotation of the bike in a 2D environment, and $u$, $v$, and $\omega$ are the longitudinal, lateral, and angular velocities.
We set $a \in [-5,2]$, $\delta \in [-\frac{\pi}{4}, \frac{\pi}{4}]$, and the rectangular robot sizes $10 \times 20$.
The values for the bicycle model specific simulation parameters $m, l_f, l_r, k_f, k_r, I_z$ were taken from Table 1 in~\cite{ge_numerically_2020}.

\begin{table*}[!ht]
    \centering
    \renewcommand{\tabcolsep}{2.5pt}
    \renewcommand{\arraystretch}{0.6}
    \footnotesize
    \caption{Results of KRRF and LazyTSP on the Car-like model and SFF* for the Dubins car (average $\pm$ std. deviation)\vspace{-0.8em}}
    \label{tab:results_car}
    \newcolumntype{C}{>{\hspace*{-4pt}\scriptsize}c<{\hspace*{-4pt}}}
    \newcolumntype{R}{>{\scriptsize}r<{}}
    \begin{tabular}{l l  c r rCR rCR   c   r rCR rCR   c    r rCR rCR}
        \toprule
        \multicolumn{2}{c}{Number of targets:}  &           & \multicolumn{7}{c}{5} &                     & \multicolumn{7}{c}{10}            &                                   & \multicolumn{7}{c}{20}                                                                                                                                                                                                                                                                                                                                     \\
        \cmidrule{1-2} \cmidrule{4-10} \cmidrule{12-18} \cmidrule{20-26}
                                           &           &                       & \textbf{Success}    & \multicolumn{3}{c}{\textbf{Cost}} & \multicolumn{3}{c}{\textbf{Time}} &                        & \textbf{Success}    & \multicolumn{3}{c}{\textbf{Cost}} & \multicolumn{3}{c}{\textbf{Time}} &  & \textbf{Success}    & \multicolumn{3}{c}{\textbf{Cost}} & \multicolumn{3}{c}{\textbf{Time}}                                                                                                                                      \\
        Map                                & \ Alg.    &                       & \textbf{rate} \hfil & \multicolumn{3}{c}{$\times 10^3$} & \multicolumn{3}{c}{seconds}       &                        & \textbf{rate} \hfil & \multicolumn{3}{c}{$\times 10^3$} & \multicolumn{3}{c}{seconds}       &  & \textbf{rate} \hfil & \multicolumn{3}{c}{$\times 10^3$} & \multicolumn{3}{c}{seconds}                                                                                                                                            \\
        \midrule
        \multirow{3}{*}{\textbf{Potholes}} & \ KRRF    &                       & $100 \%$            & $\textbf{2.80}$                   & $\pm$                             & $0.47$                 & $11.07$             & $\pm$                             & $40.29$                           &  & $100 \%$            & $\textbf{4.73}$                   & $\pm$                             & $0.83$ & $23.88$          & $\pm$ & $49.64$ &  & $93 \%$  & $\textbf{7.00}$  & $\pm$ & $0.98$ & $59.78$          & $\pm$ & $94.25$ \\
                                           & \ LazyTSP &                       & $100 \%$            & $3.29$                            & $\pm$                             & $0.90$                 & $5.53$              & $\pm$                             & $4.75$                            &  & $100 \%$            & $5.46$                            & $\pm$                             & $1.08$ & $33.32$          & $\pm$ & $14.03$ &  & $98 \%$  & $10.36$          & $\pm$ & $1.79$ & $274.30$         & $\pm$ & $89.60$ \\
                                           & \ SFF*    &                       & $100 \%$            & $7.38$                            & $\pm$                             & $0.48$                 & $\textbf{1.89}$     & $\pm$                             & $0.09$                            &  & $100 \%$            & $9.05$                            & $\pm$                             & $1.06$ & $\textbf{2.62}$  & $\pm$ & $0.18$  &  & $100 \%$ & $10.98$          & $\pm$ & $0.93$ & $\textbf{4.01}$  & $\pm$ & $0.40$  \\
        \midrule
        \multirow{3}{*}{\textbf{Floor}}    & \ KRRF    &                       & $96 \%$             & $\textbf{14.14}$                  & $\pm$                             & $1.66$                 & $21.08$             & $\pm$                             & $13.50$                           &  & $100 \%$            & $\textbf{17.86}$                  & $\pm$                             & $2.05$ & $35.36$          & $\pm$ & $25.22$ &  & $100 \%$ & $\textbf{24.28}$ & $\pm$ & $1.49$ & $83.49$          & $\pm$ & $62.26$ \\
                                           & \ LazyTSP &                       & $33 \%$             & $15.55$                           & $\pm$                             & $2.25$                 & $258.87$            & $\pm$                             & $108.13$                          &  & $0 \%$              & $-$                               & $\pm$                             & $-$    & $-$              & $\pm$ & $-$     &  & $0 \%$   & $-$              & $\pm$ & $-$    & $-$              & $\pm$ & $-$     \\
                                           & \ SFF*    &                       & $100 \%$            & $23.91$                           & $\pm$                             & $1.47$                 & $\textbf{20.19}$    & $\pm$                             & $1.19$                            &  & $100 \%$            & $35.07$                           & $\pm$                             & $4.38$ & $\textbf{28.09}$ & $\pm$ & $1.20$  &  & $100 \%$ & $30.10$          & $\pm$ & $3.14$ & $\textbf{33.09}$ & $\pm$ & $0.87$  \\
        \midrule
        \multirow{3}{*}{\textbf{Bugtrap}}  & \ KRRF    &                       & $100 \%$            & $\textbf{6.76}$                   & $\pm$                             & $0.76$                 & $\textbf{2.62}$     & $\pm$                             & $10.16$                           &  & $100 \%$            & $\textbf{6.41}$                   & $\pm$                             & $0.58$ & $\textbf{4.33}$  & $\pm$ & $16.41$ &  & $100 \%$ & $\textbf{7.07}$  & $\pm$ & $0.83$ & $20.70$          & $\pm$ & $33.35$ \\
                                           & \ LazyTSP &                       & $100 \%$            & $6.88$                            & $\pm$                             & $0.95$                 & $6.20$              & $\pm$                             & $3.91$                            &  & $100 \%$            & $7.27$                            & $\pm$                             & $1.28$ & $21.58$          & $\pm$ & $9.14$  &  & $100 \%$ & $10.66$          & $\pm$ & $2.31$ & $125.00$         & $\pm$ & $43.74$ \\
                                           & \ SFF*    &                       & $100 \%$            & $7.08$                            & $\pm$                             & $0.48$                 & $4.02$              & $\pm$                             & $0.27$                            &  & $100 \%$            & $11.47$                           & $\pm$                             & $1.54$ & $4.91$           & $\pm$ & $0.27$  &  & $100 \%$ & $13.87$          & $\pm$ & $1.10$ & $\textbf{6.37}$  & $\pm$ & $0.39$  \\
        \midrule
        \multirow{3}{*}{\textbf{DoubleBT}} & \ KRRF    &                       & $99 \%$             & $\textbf{16.65}$                  & $\pm$                             & $1.63$                 & $\textbf{16.16}$    & $\pm$                             & $36.39$                           &  & $100 \%$            & $\textbf{25.67}$                  & $\pm$                             & $1.74$ & $\textbf{6.74}$  & $\pm$ & $5.54$  &  & $100 \%$ & $\textbf{27.98}$ & $\pm$ & $2.28$ & $\textbf{43.14}$ & $\pm$ & $97.22$ \\
                                           & \ LazyTSP &                       & $100 \%$            & $19.20$                           & $\pm$                             & $2.02$                 & $33.63$             & $\pm$                             & $9.69$                            &  & $100 \%$            & $26.06$                           & $\pm$                             & $2.30$ & $156.92$         & $\pm$ & $34.24$ &  & $3 \%$   & $33.34$          & $\pm$ & $1.72$ & $282.16$         & $\pm$ & $32.56$ \\
                                           & \ SFF*    &                       & $100 \%$            & $29.42$                           & $\pm$                             & $1.81$                 & $38.81$             & $\pm$                             & $2.04$                            &  & $100 \%$            & $29.42$                           & $\pm$                             & $1.81$ & $38.81$          & $\pm$ & $2.04$  &  & $100 \%$ & $39.24$          & $\pm$ & $3.63$ & $57.57$          & $\pm$ & $1.74$  \\
        \bottomrule
    \end{tabular}
    \vspace{-2.5em}
\end{table*}

\subsection{Experiment setup} \label{sec:Experiment setup}

All algorithms were tested on four maps, and for 5, 10, and 20 targets.
The maps with corresponding sizes in map units are Potholes ($900\times900$), Floor ($2100\times2300$), Bugtrap ($1300\times1000$) and Double Bugtrap (DoubleBT) ($3600\times2000$) (shown in Fig.~\ref{fig::map_intro}).

The parameters of KRRF are described in Sec.~\ref{sec:Discussion in method}.
For the nearest neighbor search, the FLANN library~\cite{muja2009flann} was selected.
Since there are currently no algorithms except ours suitable for a general robot dynamic model, we compared our proposed algorithm with the LazyTSP modified for multi-goal planning~\cite{englot2013three}.
Moreover, since the SFF* algorithm~\cite{janos_randomized_2022} can only handle the Dubins vehicle, the comparison to SFF* was made only using this model.
Lazy-TSP was modified to create continuous trajectories through all targets.
This is significantly slower than only running a simple LazyTSP returning a collection of discontinuous trajectories in the targets, however, it is necessary for computing a full feasible trajectory.
This algorithm was run with similar parameters to KRRF, only the number of iterations to plan one target-to-target subtrajectory was increased to $3\cdot10^{4}$.

The parameters of SFF* from~\cite{janos_randomized_2022} were set 
to mimic the dynamics of the Car-like robot:
number of iterations $10^6$, collision checking at resolution $5$ (map units), goal-bias $0.95$, dubins-radius $10$ (map units), and edge time $4$~s.
The other parameters were kept at their default values.

\subsection{Car-like model experiments} \label{sec:Car-like experiments}

The first experiment compares KRRF with LazyTSP for the Car-like model and SFF* with a similar Dubins vehicle model.
Algorithms are compared using the path length rather than the trajectory duration.
This choice allows comparison with SFF* for the Dubins vehicle model path calculation.

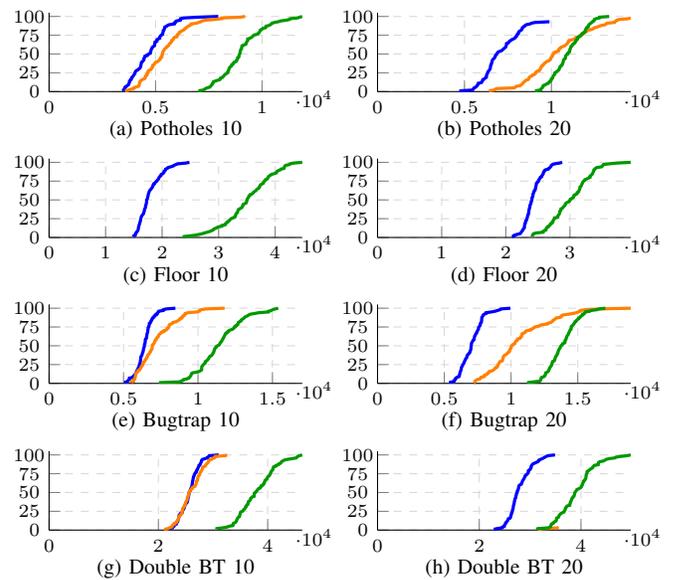
\begin{figure}[!t]
    \centering
    \input{figs/tikz_hist}
    \vspace*{-15pt}
    \caption{
    Cumulative distribution of Trajectory length of all algorithms on all maps with 10 and 20 targets. Probability (vertical axis) is in \%, while trajectory cost (horizontal) in map units.
    \textcolor{blue}{\textbf{---}}~KRRF\hspace{1em}
    \textcolor{orange}{\textbf{---}}~LazyTSP\hspace{1em}
    \textcolor{green!60!black}{\textbf{---}}~SFF*
    }
    \vspace*{-14pt}
    \label{fig::cummulative}
\end{figure}

The results over 100 trials are summarized in Table~\ref{tab:results_car}, the trajectory length cumulative distributions in Fig.~\ref{fig::cummulative}, and  the selected final paths are depicted in Fig.~\ref{fig::map_intro}.
The success rate denotes the percentage of trials in which a solution was successfully found in under 10 minutes.
Our method has the shortest trajectories (column \textbf{Cost}) in all scenarios --- in some cases, about two times shorter than SFF*.
The trajectory length distributions in Fig.~\ref{fig::cummulative} show that the results of KRRF are very consistent, achieving the shortest lengths in all maps.
SFF* is faster in most of the experiments as it uses an analytical solution to BVP. In contrast, other methods do not use this solution and use a (time-consuming) forward simulation of the motion model.
However, SFF* works only for Car-like robots and not for a general kinodynamic system.
While SFF* is faster than other methods, its runtime varies in the environments, which is caused
by their different complexity and the number of narrow passages 
(Potholes have no narrow passages, Bugtrap has only one, and Floor and DoubleBT have multiple narrow passages).
These narrow passages need to be traversed by the SFF* trees, which increases its runtime.

Lazy-TSP is about five times slower than KRRF for larger instances with 20 targets, while for the Floor map, it could only solve the 5 target instances with 33\% success rate.
However, the length of the trajectories is, on average, better than SFF* in all scenarios but worse than of the proposed KRRF.
\Review{The runtime of KRRF increases approximately linearly with the number of targets and not as dramatically as in the case of the LazyTSP.}





\begin{table*}[t]
    \centering
    \renewcommand{\tabcolsep}{2.1pt}
    \renewcommand{\arraystretch}{0.6}
    \footnotesize
    \caption{Results of KRRF and LazyTSP on Differential drive model and Bike model (average $\pm$ std. deviation)\vspace{-0.8em}}
    \label{tab:results_bike_diff}

    \newcolumntype{C}{>{\hspace*{-3.5pt}\scriptsize}c<{\hspace*{-3.5pt}}}
    \newcolumntype{R}{>{\scriptsize}r<{}}
    \begin{tabular}{l l l c r rCR rCR   c   r rCR rCR   c    r rCR rCR}
        \toprule
        \multicolumn{3}{c}{Number of targets:}  & \multicolumn{7}{c}{5} &      & \multicolumn{7}{c}{10} &                     & \multicolumn{7}{c}{20}                                                                                                                                                                                                                                                                                                                                                                                               \\
        \cmidrule{1-3} \cmidrule{5-11} \cmidrule{13-19} \cmidrule{21-27}
                                           &                       &      &                        & \textbf{Success}    & \multicolumn{3}{c}{\textbf{Cost}} & \multicolumn{3}{c}{\textbf{Time}} &        & \textbf{Success}    & \multicolumn{3}{c}{\textbf{Cost}} & \multicolumn{3}{c}{\textbf{Time}} &  & \textbf{Success}    & \multicolumn{3}{c}{\textbf{Cost}} & \multicolumn{3}{c}{\textbf{Time}}                                                                                                                                        \\
        Map                                & Alg.                  & Model &                        & \textbf{rate} \hfil & \multicolumn{3}{c}{$\times 10^3$} & \multicolumn{3}{c}{seconds}       &        & \textbf{rate} \hfil & \multicolumn{3}{c}{$\times 10^3$} & \multicolumn{3}{c}{seconds}       &  & \textbf{rate} \hfil & \multicolumn{3}{c}{$\times 10^3$} & \multicolumn{3}{c}{seconds}                                                                                                                                              \\
        \midrule
        \multirow{4}{*}{\textbf{Potholes}} & KRRF                  & diff &                        & $100 \%$            & $\textbf{2.42}$                   & $\pm$                             & $0.30$ & $\textbf{0.61}$     & $\pm$                             & $0.60$                            &  & $100 \%$            & $\textbf{4.07}$                   & $\pm$                             & $0.63$ & $\textbf{2.32}$   & $\pm$ & $1.05$  &  & $100 \%$ & $\textbf{5.21}$  & $\pm$ & $0.47$ & $\textbf{7.11}$   & $\pm$ & $6.21$  \\
                                           & LazyTSP               & diff &                        & $100 \%$            & $2.52$                            & $\pm$                             & $0.70$ & $4.67$              & $\pm$                             & $1.71$                            &  & $100 \%$            & $4.45$                            & $\pm$                             & $0.76$ & $22.37$           & $\pm$ & $7.40$  &  & $100 \%$ & $7.58$           & $\pm$ & $1.49$ & $48.56$           & $\pm$ & $18.89$ \\
                                           & KRRF                  & bike &                        & $100 \%$            & $\textbf{2.15}$                   & $\pm$                             & $0.19$ & $\textbf{0.40}$     & $\pm$                             & $0.21$                            &  & $100 \%$            & $\textbf{3.77}$                   & $\pm$                             & $0.30$ & $\textbf{2.02}$   & $\pm$ & $3.91$  &  & $99 \%$  & $\textbf{5.50}$  & $\pm$ & $0.42$ & $\textbf{8.33}$   & $\pm$ & $11.50$ \\
                                           & LazyTSP               & bike &                        & $100 \%$            & $2.59$                            & $\pm$                             & $0.43$ & $2.58$              & $\pm$                             & $1.12$                            &  & $100 \%$            & $4.74$                            & $\pm$                             & $0.61$ & $11.27$           & $\pm$ & $3.24$  &  & $100 \%$ & $7.42$           & $\pm$ & $0.84$ & $72.11$           & $\pm$ & $20.36$ \\
        \midrule
        \multirow{4}{*}{\textbf{Floor}}    & KRRF                  & diff &                        & $96 \%$             & $\textbf{12.69}$                  & $\pm$                             & $1.22$ & $\textbf{16.16}$    & $\pm$                             & $10.95$                           &  & $99 \%$             & $\textbf{16.14}$                  & $\pm$                             & $1.97$ & $\textbf{24.61}$  & $\pm$ & $10.15$ &  & $100 \%$ & $\textbf{21.26}$ & $\pm$ & $1.83$ & $\textbf{49.35}$  & $\pm$ & $15.62$ \\
                                           & LazyTSP               & diff &                        & $3 \%$              & $13.81$                           & $\pm$                             & $1.42$ & $247.45$            & $\pm$                             & $38.56$                           &  & $0 \%$              & $-$                               & $\pm$                             & $-$    & $-$               & $\pm$ & $-$     &  & $0 \%$   & $-$              & $\pm$ & $-$    & $-$               & $\pm$ & $-$     \\
                                           & KRRF                  & bike &                        & $100 \%$            & $\textbf{15.83}$                  & $\pm$                             & $1.57$ & $\textbf{37.60}$    & $\pm$                             & $28.60$                           &  & $100 \%$            & $\textbf{18.85}$                  & $\pm$                             & $1.65$ & $\textbf{45.54}$  & $\pm$ & $24.59$ &  & $99 \%$  & $\textbf{24.72}$ & $\pm$ & $1.16$ & $\textbf{167.61}$ & $\pm$ & $54.08$ \\
                                           & LazyTSP               & bike &                        & $97 \%$             & $17.23$                           & $\pm$                             & $1.65$ & $234.30$            & $\pm$                             & $111.70$                          &  & $15 \%$             & $21.64$                           & $\pm$                             & $2.51$ & $493.49$          & $\pm$ & $68.29$ &  & $0 \%$   & $-$              & $\pm$ & $-$    & $-$               & $\pm$ & $-$     \\
        \midrule
        \multirow{4}{*}{\textbf{Bugtrap}}  & KRRF                  & diff &                        & $100 \%$            & $\textbf{6.33}$                   & $\pm$                             & $0.63$ & $\textbf{1.64}$     & $\pm$                             & $1.38$                            &  & $100 \%$            & $\textbf{5.62}$                   & $\pm$                             & $0.47$ & $\textbf{2.04}$   & $\pm$ & $0.42$  &  & $100 \%$ & $\textbf{5.66}$  & $\pm$ & $0.61$ & $\textbf{5.96}$   & $\pm$ & $1.85$  \\
                                           & LazyTSP               & diff &                        & $100 \%$            & $6.42$                            & $\pm$                             & $0.85$ & $10.14$             & $\pm$                             & $1.61$                            &  & $100 \%$            & $6.51$                            & $\pm$                             & $1.35$ & $31.62$           & $\pm$ & $8.51$  &  & $100 \%$ & $8.44$           & $\pm$ & $2.07$ & $78.29$           & $\pm$ & $26.87$ \\
                                           & KRRF                  & bike &                        & $100 \%$            & $\textbf{6.42}$                   & $\pm$                             & $0.49$ & $\textbf{2.14}$     & $\pm$                             & $1.50$                            &  & $100 \%$            & $\textbf{6.36}$                   & $\pm$                             & $0.36$ & $\textbf{3.79}$   & $\pm$ & $1.68$  &  & $100 \%$ & $\textbf{6.43}$  & $\pm$ & $0.35$ & $\textbf{6.88}$   & $\pm$ & $1.22$  \\
                                           & LazyTSP               & bike &                        & $100 \%$            & $6.90$                            & $\pm$                             & $0.52$ & $6.13$              & $\pm$                             & $1.05$                            &  & $100 \%$            & $7.37$                            & $\pm$                             & $0.81$ & $23.05$           & $\pm$ & $3.57$  &  & $100 \%$ & $7.97$           & $\pm$ & $0.70$ & $68.26$           & $\pm$ & $14.49$ \\
        \midrule
        \multirow{4}{*}{\textbf{DoubleBT}} & KRRF                  & diff &                        & $76 \%$             & $\textbf{15.15}$                  & $\pm$                             & $1.22$ & $\textbf{56.77}$    & $\pm$                             & $51.04$                           &  & $100 \%$            & $\textbf{23.95}$                  & $\pm$                             & $2.00$ & $\textbf{30.00}$  & $\pm$ & $18.81$ &  & $100 \%$ & $\textbf{24.43}$ & $\pm$ & $1.75$ & $\textbf{36.61}$  & $\pm$ & $12.95$ \\
                                           & LazyTSP               & diff &                        & $23 \%$             & $16.17$                           & $\pm$                             & $1.50$ & $185.62$            & $\pm$                             & $70.82$                           &  & $0 \%$              & $-$                               & $\pm$                             & $-$    & $-$               & $\pm$ & $-$     &  & $0 \%$   & $-$              & $\pm$ & $-$    & $-$               & $\pm$ & $-$     \\
                                           & KRRF                  & bike &                        & $62 \%$             & $\textbf{19.82}$                  & $\pm$                             & $1.90$ & $\textbf{120.73}$   & $\pm$                             & $93.12$                           &  & $95 \%$             & $\textbf{27.53}$                  & $\pm$                             & $1.47$ & $\textbf{121.71}$ & $\pm$ & $51.00$ &  & $100 \%$ & $\textbf{29.39}$ & $\pm$ & $1.81$ & $\textbf{141.35}$ & $\pm$ & $69.75$ \\
                                           & LazyTSP               & bike &                        & $16 \%$             & $21.97$                           & $\pm$                             & $1.22$ & $183.93$            & $\pm$                             & $113.64$                          &  & $0 \%$              & $-$                               & $\pm$                             & $-$    & $-$               & $\pm$ & $-$     &  & $0 \%$   & $-$              & $\pm$ & $-$    & $-$               & $\pm$ & $-$     \\
        \bottomrule
    \end{tabular}
    \vspace{-2em}
\end{table*}

\subsection{Diff-drive and Bike-like model experiments} \label{sec:Diff-drive and Bike-like experiments}

We also compare the performance of KRRF and LazyTSP on the other two dynamical models: Diff-drive and Bike-like.
In Table~\ref{tab:results_bike_diff} the clear dominance of KRRF is evident in both the trajectory length and the computation time.
Low success rates of LazyTSP for DoubleBT map show that it can not successfully find a solution for almost any instance for the two robot models.
This showcases the benefits of the proposed method being able to find effectively a multi-goal plan with the highest quality for an arbitrary kinodynamic motion model.



\section{Conclusion}

This paper focused on multi-goal motion planning while respecting kinodynamic robot model.
We proposed the Kinodynamic Rapidly-exploring Random Forest (KRRF) planner, which computes a continuous trajectory over all the targets with a low cost.
KRRF grows trees from each target in the configuration space to determine trajectories between targets. 
Respective trajectory costs are used in a TSP-based solver to compute the visit order.
Finally, an RRT-based planner of the final multi-goal trajectory is guided along the trajectories in the TSP order.
The conducted experiments show that our method provides higher-quality trajectories in a shorter computational time compared to other state-of-the-art methods.
Our method finds high-quality trajectories for any kinodynamic motion model without a significant increase in computational time as the number of targets grows.

\bibliographystyle{IEEETran}
\balance
\bibliography{paper}


\end{document}

%% file: figs/tikz_hist.tex
\begin{tikzpicture}
    \pgfplotsset{
        width=0.38\linewidth,
        height=1.05cm,
        grid=major,
        major grid style={dashed, gray!30},
        ticklabel style = {font=\scriptsize},
        legend style={font=\scriptsize},
        title style={
                font=\footnotesize,
                at={(0.5,-0.45)},
                anchor=north,
            },
        scale only axis=true,
        x tick label style={anchor=north, yshift=0.0cm},
        every x tick scale label/.style={
                at={(0.96,-0.2)},xshift=1pt,anchor=south west,inner sep=0pt
            },
        y tick label style={text width=1em,align=right},
        ylabel style={
                yshift=-5.0mm,
            },
        ymin=0,
        ymax=105,
        ytick={0,25,50,75,100},
        axis lines*=left,
        every axis plot/.append style={line width=1.2pt},
        legend pos=north west,
    }

    \begin{axis}[
            name=potholes10,
            title=(a) Potholes 10,
            xmin=0,
            xmax=11887.3,
        ]
        \addplot[color=blue]
        coordinates {(3434.36,1.0)(3508.65,2.0)(3522.18,3.0)(3540.89,4.0)(3565.38,5.0)(3567.7,6.0)(3630.04,7.000000000000001)(3675.72,8.0)(3727.56,9.0)(3736.75,10.0)(3753.82,11.0)(3775.25,12.0)(3775.59,13.0)(3777.41,14.000000000000002)(3807.68,15.0)(3854.17,16.0)(3854.78,17.0)(3879.53,18.0)(3887.96,19.0)(3937.5,20.0)(3955.75,21.0)(3978.56,22.0)(4004.25,23.0)(4030.51,24.0)(4033.04,25.0)(4047.21,26.0)(4063.76,27.0)(4098.98,28.000000000000004)(4179.09,28.999999999999996)(4205.17,30.0)(4258.75,31.0)(4284.72,32.0)(4294.4,33.0)(4305.87,34.0)(4310.97,35.0)(4317.05,36.0)(4323.73,37.0)(4336.07,38.0)(4338.7,39.0)(4389.92,40.0)(4414.82,41.0)(4448.28,42.0)(4449.54,43.0)(4450.47,44.0)(4480.98,45.0)(4488.06,46.0)(4496.48,47.0)(4610.49,48.0)(4616.47,49.0)(4677.71,50.0)(4754.31,51.0)(4761.33,52.0)(4767.44,53.0)(4780.43,54.0)(4784.71,55.00000000000001)(4801.84,56.00000000000001)(4808.73,56.99999999999999)(4838.87,57.99999999999999)(4848.37,59.0)(4895.95,60.0)(4950.94,61.0)(4959.47,62.0)(4961.26,63.0)(4963.71,64.0)(4973.31,65.0)(4976.17,66.0)(5063.82,67.0)(5064.17,68.0)(5066.04,69.0)(5111.55,70.0)(5197.38,71.0)(5197.73,72.0)(5202.7,73.0)(5228.14,74.0)(5268.15,75.0)(5324.44,76.0)(5326.07,77.0)(5345.49,78.0)(5349.49,79.0)(5357.47,80.0)(5357.82,81.0)(5395.58,82.0)(5406.41,83.0)(5421.62,84.0)(5427.71,85.0)(5437.08,86.0)(5522.57,87.0)(5560.46,88.0)(5637.05,89.0)(5873.63,90.0)(5874.22,91.0)(5930.76,92.0)(5931.48,93.0)(5979.12,94.0)(6164.53,95.0)(6174.33,96.0)(6174.5,97.0)(6331.17,98.0)(6968.06,99.0)(7946.61,100.0)};
        \addplot[color=orange]
        coordinates {(3568.56,1.0)(3728.92,2.0)(3828.29,3.0)(3973.79,4.0)(3986.84,5.0)(4069.12,6.0)(4194.35,7.000000000000001)(4222.02,8.0)(4231.5,9.0)(4241.66,10.0)(4243.06,11.0)(4258.61,12.0)(4259.36,13.0)(4294.72,14.000000000000002)(4329.39,15.0)(4376.22,16.0)(4476.25,17.0)(4510.25,18.0)(4517.47,19.0)(4520.49,20.0)(4537.81,21.0)(4600.36,22.0)(4604.54,23.0)(4604.55,24.0)(4637.53,25.0)(4637.98,26.0)(4650.87,27.0)(4672.46,28.000000000000004)(4767.69,28.999999999999996)(4789.08,30.0)(4827.82,31.0)(4835.41,32.0)(4863.87,33.0)(4865.39,34.0)(4866.34,35.0)(4932.48,36.0)(4953.36,37.0)(4977.44,38.0)(4999.52,39.0)(5105.33,40.0)(5138.13,41.0)(5180.33,42.0)(5216.73,43.0)(5221.36,44.0)(5231.02,45.0)(5292.18,46.0)(5295.61,47.0)(5299.41,48.0)(5318.5,49.0)(5320.21,50.0)(5327.52,51.0)(5331.73,52.0)(5354.82,53.0)(5366.28,54.0)(5394.62,55.00000000000001)(5412.1,56.00000000000001)(5443.62,56.99999999999999)(5477.03,57.99999999999999)(5522.5,59.0)(5532.02,60.0)(5600.22,61.0)(5621.25,62.0)(5645.42,63.0)(5670.42,64.0)(5711.85,65.0)(5739.17,66.0)(5749.66,67.0)(5756.47,68.0)(5826.92,69.0)(5843.18,70.0)(5850.82,71.0)(5870.38,72.0)(6002.76,73.0)(6012.64,74.0)(6013.09,75.0)(6034.72,76.0)(6085.34,77.0)(6116.65,78.0)(6120.78,79.0)(6187.18,80.0)(6255.76,81.0)(6318.36,82.0)(6367.21,83.0)(6385.09,84.0)(6439.47,85.0)(6484.08,86.0)(6520.89,87.0)(6615.25,88.0)(6654.09,89.0)(6664.41,90.0)(6907.71,91.0)(7028.61,92.0)(7127.15,93.0)(7174.8,94.0)(7354.08,95.0)(7763.88,96.0)(8100.29,97.0)(8107.95,98.0)(8972.12,99.0)(9206.1,100.0)};
        \addplot[color=green!60!black]
        coordinates {(7009.89,1.0)(7224.99,2.0)(7297.05,3.0)(7341.06,4.0)(7419.97,5.0)(7552.83,6.0)(7574.51,7.000000000000001)(7613.24,8.0)(7628.11,9.0)(7672.39,10.0)(7674.49,11.0)(7707.03,12.0)(7714.73,13.0)(7803.64,14.000000000000002)(7907.78,15.0)(8023.04,16.0)(8039.99,17.0)(8049.89,18.0)(8149.14,19.0)(8172.71,20.0)(8204.69,21.0)(8234.07,22.0)(8275.41,23.0)(8329.51,24.0)(8388.96,25.0)(8396.2,26.0)(8396.62,27.0)(8425.53,28.000000000000004)(8442.04,28.999999999999996)(8443.24,30.0)(8460.61,31.0)(8474.76,32.0)(8512.75,33.0)(8565.2,34.0)(8574.61,35.0)(8580.67,36.0)(8652.82,37.0)(8659.55,38.0)(8683.14,39.0)(8780.18,40.0)(8804.92,41.0)(8805.71,42.0)(8840.46,43.0)(8852.73,44.0)(8853.06,45.0)(8908.28,46.0)(8919.84,47.0)(8919.85,48.0)(8929.89,49.0)(8943.45,50.0)(8966.34,51.0)(8969.71,52.0)(8972.88,53.0)(8977.66,54.0)(8995.72,55.00000000000001)(9001.3,56.00000000000001)(9035.78,56.99999999999999)(9042.22,57.99999999999999)(9045.65,59.0)(9050.07,60.0)(9060.03,61.0)(9147.18,62.0)(9241.31,63.0)(9251.49,64.0)(9261.96,65.0)(9275.82,66.0)(9306.34,67.0)(9321.69,68.0)(9323.84,69.0)(9325.19,70.0)(9428.29,71.0)(9493.14,72.0)(9543.61,73.0)(9571.68,74.0)(9608.77,75.0)(9723.5,76.0)(9821.28,77.0)(9839.74,78.0)(9878.11,79.0)(9891.76,80.0)(9918.59,81.0)(9924.14,82.0)(9996.04,83.0)(10032.6,84.0)(10164.8,85.0)(10185.1,86.0)(10188.8,87.0)(10319.1,88.0)(10393.6,89.0)(10436.0,90.0)(10467.7,91.0)(10584.1,92.0)(10772.0,93.0)(10872.2,94.0)(11132.9,95.0)(11288.8,96.0)(11317.0,97.0)(11659.3,98.0)(11774.3,99.0)(11887.3,100.0)};
    \end{axis}
    \begin{axis}[
            name=floor10,
            title=(c) Floor 10,
            at=(potholes10.below south west), anchor=above north west,
            xmin=0,
            xmax=44793.9,
        ]
        \addplot[color=blue]
        coordinates {(14862.9,1.0)(14940.9,2.0)(15254.2,3.0)(15285.9,4.0)(15290.8,5.0)(15413.2,6.0)(15443.9,7.000000000000001)(15587.5,8.0)(15700.5,9.0)(15742.0,10.0)(15788.9,11.0)(15797.8,12.0)(15803.2,13.0)(15865.3,14.000000000000002)(15874.2,15.0)(15904.2,16.0)(15950.9,17.0)(16017.4,18.0)(16039.2,19.0)(16067.3,20.0)(16113.1,21.0)(16161.9,22.0)(16183.0,23.0)(16206.0,24.0)(16227.2,25.0)(16265.1,26.0)(16303.3,27.0)(16339.9,28.000000000000004)(16447.0,28.999999999999996)(16510.2,30.0)(16643.0,31.0)(16730.3,32.0)(16741.9,33.0)(16790.3,34.0)(16847.5,35.0)(16863.7,36.0)(16874.5,37.0)(16924.5,38.0)(17004.2,39.0)(17056.9,40.0)(17063.4,41.0)(17070.3,42.0)(17145.7,43.0)(17146.3,44.0)(17200.7,45.0)(17232.0,46.0)(17240.9,47.0)(17264.3,48.0)(17264.6,49.0)(17265.9,50.0)(17352.1,51.0)(17418.0,52.0)(17438.6,53.0)(17440.6,54.0)(17515.2,55.00000000000001)(17516.4,56.00000000000001)(17555.8,56.99999999999999)(17571.5,57.99999999999999)(17631.2,59.0)(17681.1,60.0)(17710.7,61.0)(17771.1,62.0)(17868.1,63.0)(18001.5,64.0)(18086.5,65.0)(18201.7,66.0)(18280.2,67.0)(18359.0,68.0)(18372.0,69.0)(18419.1,70.0)(18602.4,71.0)(18605.4,72.0)(18778.9,73.0)(18881.9,74.0)(19070.1,75.0)(19123.4,76.0)(19282.2,77.0)(19374.0,78.0)(19399.5,79.0)(19486.2,80.0)(19567.1,81.0)(19772.5,82.0)(19780.8,83.0)(19852.9,84.0)(20156.8,85.0)(20273.3,86.0)(20317.6,87.0)(20338.1,88.0)(20536.1,89.0)(20606.6,90.0)(20623.1,91.0)(20938.4,92.0)(21034.4,93.0)(21494.9,94.0)(21951.5,95.0)(22084.3,96.0)(22428.7,97.0)(22747.3,98.0)(24168.8,99.0)(24842.4,100.0)};
        \addplot[color=orange]
        coordinates {};
        \addplot[color=green!60!black]
        coordinates {(23633.7,1.0)(24658.9,2.0)(25835.6,3.0)(26478.6,4.0)(27108.8,5.0)(27468.5,6.0)(27976.9,7.000000000000001)(28413.1,8.0)(28734.5,9.0)(29033.5,10.0)(29099.4,11.0)(29382.1,12.0)(29675.1,13.0)(29908.2,14.000000000000002)(30555.9,15.0)(30600.1,16.0)(31181.2,17.0)(31221.8,18.0)(31294.0,19.0)(31533.4,20.0)(31567.2,21.0)(31837.3,22.0)(31935.6,23.0)(32073.6,24.0)(32187.4,25.0)(32520.1,26.0)(32866.5,27.0)(32900.1,28.000000000000004)(33002.4,28.999999999999996)(33023.2,30.0)(33038.4,31.0)(33112.1,32.0)(33311.2,33.0)(33343.7,34.0)(33380.3,35.0)(33436.1,36.0)(33461.0,37.0)(33673.0,38.0)(33962.6,39.0)(34118.7,40.0)(34175.9,41.0)(34223.1,42.0)(34336.4,43.0)(34339.2,44.0)(34639.8,45.0)(34880.1,46.0)(34900.4,47.0)(34952.0,48.0)(35015.3,49.0)(35017.1,50.0)(35062.6,51.0)(35122.2,52.0)(35255.9,53.0)(35312.3,54.0)(35342.2,55.00000000000001)(35528.8,56.00000000000001)(35553.1,56.99999999999999)(35915.1,57.99999999999999)(36083.6,59.0)(36136.1,60.0)(36307.1,61.0)(36543.7,62.0)(36660.0,63.0)(36749.7,64.0)(36946.5,65.0)(37167.2,66.0)(37175.0,67.0)(37285.4,68.0)(37331.4,69.0)(37397.9,70.0)(37627.6,71.0)(37881.0,72.0)(37886.4,73.0)(38034.2,74.0)(38079.2,75.0)(38234.1,76.0)(38309.2,77.0)(38638.8,78.0)(38684.5,79.0)(38749.0,80.0)(38805.8,81.0)(38991.1,82.0)(39194.3,83.0)(39344.0,84.0)(39871.7,85.0)(40224.2,86.0)(40246.6,87.0)(40342.3,88.0)(40550.1,89.0)(40916.1,90.0)(41037.6,91.0)(41583.4,92.0)(41621.9,93.0)(41656.7,94.0)(41692.4,95.0)(41876.1,96.0)(42508.2,97.0)(42659.0,98.0)(43250.0,99.0)(44793.9,100.0)};
    \end{axis}
    \begin{axis}[
            name=bugtrap10,
            title=(e) Bugtrap 10,
            at=(floor10.below south west), anchor=above north west,
            xmin=0,
            xmax=17000,
        ]
        \addplot[color=blue]
        coordinates {(5037.46,1.0)(5184.88,2.0)(5290.72,3.0)(5327.83,4.0)(5386.55,5.0)(5388.31,6.0)(5393.82,7.000000000000001)(5621.77,8.0)(5651.75,9.0)(5675.7,10.0)(5756.14,11.0)(5775.98,12.0)(5808.14,13.0)(5820.3,14.000000000000002)(5868.16,15.0)(5881.94,16.0)(5910.44,17.0)(5921.97,18.0)(5960.11,19.0)(5996.52,20.0)(5997.37,21.0)(6030.87,22.0)(6040.55,23.0)(6049.53,24.0)(6061.98,25.0)(6091.72,26.0)(6109.68,27.0)(6110.46,28.000000000000004)(6134.96,28.999999999999996)(6136.38,30.0)(6137.38,31.0)(6148.5,32.0)(6154.69,33.0)(6156.12,34.0)(6156.91,35.0)(6186.66,36.0)(6226.87,37.0)(6240.27,38.0)(6251.79,39.0)(6268.31,40.0)(6281.05,41.0)(6308.91,42.0)(6316.89,43.0)(6333.31,44.0)(6335.64,45.0)(6343.17,46.0)(6357.56,47.0)(6386.08,48.0)(6390.63,49.0)(6399.33,50.0)(6401.71,51.0)(6407.59,52.0)(6442.5,53.0)(6445.3,54.0)(6453.44,55.00000000000001)(6458.05,56.00000000000001)(6461.71,56.99999999999999)(6461.78,57.99999999999999)(6479.52,59.0)(6540.67,60.0)(6544.78,61.0)(6549.77,62.0)(6557.95,63.0)(6569.75,64.0)(6573.8,65.0)(6574.97,66.0)(6598.88,67.0)(6618.72,68.0)(6624.01,69.0)(6632.32,70.0)(6633.61,71.0)(6648.44,72.0)(6651.15,73.0)(6651.9,74.0)(6685.88,75.0)(6704.73,76.0)(6709.89,77.0)(6729.37,78.0)(6733.36,79.0)(6735.71,80.0)(6815.48,81.0)(6829.86,82.0)(6838.23,83.0)(6893.26,84.0)(6895.14,85.0)(6933.46,86.0)(6945.43,87.0)(7044.71,88.0)(7050.14,89.0)(7073.6,90.0)(7191.09,91.0)(7192.02,92.0)(7285.54,93.0)(7286.11,94.0)(7337.16,95.0)(7362.81,96.0)(7671.27,97.0)(7680.26,98.0)(7850.02,99.0)(8487.64,100.0)};
        \addplot[color=orange]
        coordinates {(5495.47,1.0)(5504.99,2.0)(5676.36,3.0)(5697.01,4.0)(5698.83,5.0)(5753.21,6.0)(5758.32,7.000000000000001)(5781.47,8.0)(5796.43,9.0)(5801.01,10.0)(5803.08,11.0)(5821.02,12.0)(5836.32,13.0)(5838.3,14.000000000000002)(5932.09,15.0)(5970.84,16.0)(5974.48,17.0)(6083.95,18.0)(6102.99,19.0)(6144.04,20.0)(6169.81,21.0)(6189.0,22.0)(6233.62,23.0)(6275.8,24.0)(6293.78,25.0)(6306.99,26.0)(6322.35,27.0)(6329.64,28.000000000000004)(6396.22,28.999999999999996)(6437.91,30.0)(6454.84,31.0)(6475.92,32.0)(6535.65,33.0)(6557.43,34.0)(6582.99,35.0)(6594.72,36.0)(6636.89,37.0)(6639.97,38.0)(6704.61,39.0)(6737.39,40.0)(6754.49,41.0)(6810.33,42.0)(6815.38,43.0)(6858.18,44.0)(6882.44,45.0)(6913.77,46.0)(6915.4,47.0)(6921.38,48.0)(6932.68,49.0)(6934.76,50.0)(6937.53,51.0)(7005.42,52.0)(7029.56,53.0)(7062.93,54.0)(7091.73,55.00000000000001)(7098.0,56.00000000000001)(7119.25,56.99999999999999)(7181.48,57.99999999999999)(7183.5,59.0)(7229.93,60.0)(7249.37,61.0)(7330.81,62.0)(7337.46,63.0)(7354.62,64.0)(7394.75,65.0)(7456.67,66.0)(7514.19,67.0)(7533.57,68.0)(7697.99,69.0)(7699.49,70.0)(7866.49,71.0)(7910.17,72.0)(7920.64,73.0)(8019.61,74.0)(8048.07,75.0)(8051.87,76.0)(8057.16,77.0)(8082.11,78.0)(8187.63,79.0)(8280.56,80.0)(8307.3,81.0)(8532.98,82.0)(8607.85,83.0)(8701.96,84.0)(8770.05,85.0)(8783.78,86.0)(8922.52,87.0)(8926.68,88.0)(8951.6,89.0)(9062.98,90.0)(9071.64,91.0)(9139.92,92.0)(9149.25,93.0)(9328.64,94.0)(9931.77,95.0)(10082.9,96.0)(10097.7,97.0)(10206.0,98.0)(10503.4,99.0)(11787.2,100.0)};
        \addplot[color=green!60!black]
        coordinates {(7381.5,1.0)(8746.64,2.0)(8873.21,3.0)(8999.77,4.0)(9023.83,5.0)(9244.34,6.0)(9391.76,7.000000000000001)(9416.68,8.0)(9464.98,9.0)(9512.57,10.0)(9532.54,11.0)(9549.49,12.0)(9564.31,13.0)(9637.22,14.000000000000002)(9923.18,15.0)(10083.7,16.0)(10182.8,17.0)(10245.1,18.0)(10246.7,19.0)(10246.9,20.0)(10248.7,21.0)(10250.6,22.0)(10280.4,23.0)(10313.0,24.0)(10374.6,25.0)(10423.0,26.0)(10480.2,27.0)(10485.0,28.000000000000004)(10489.9,28.999999999999996)(10576.5,30.0)(10635.0,31.0)(10666.8,32.0)(10713.7,33.0)(10730.0,34.0)(10768.8,35.0)(10799.2,36.0)(10913.3,37.0)(10942.0,38.0)(10982.4,39.0)(10997.8,40.0)(11047.8,41.0)(11059.3,42.0)(11065.8,43.0)(11101.2,44.0)(11101.9,45.0)(11115.4,46.0)(11159.7,47.0)(11222.6,48.0)(11227.6,49.0)(11301.0,50.0)(11410.8,51.0)(11450.3,52.0)(11473.1,53.0)(11485.7,54.0)(11552.8,55.00000000000001)(11570.8,56.00000000000001)(11616.5,56.99999999999999)(11628.3,57.99999999999999)(11670.3,59.0)(11684.7,60.0)(11731.4,61.0)(11747.3,62.0)(11773.7,63.0)(11792.8,64.0)(11825.7,65.0)(11837.3,66.0)(11935.9,67.0)(11944.1,68.0)(12039.7,69.0)(12059.0,70.0)(12167.3,71.0)(12320.9,72.0)(12375.9,73.0)(12433.8,74.0)(12438.3,75.0)(12545.7,76.0)(12554.1,77.0)(12666.6,78.0)(12678.7,79.0)(12701.6,80.0)(12708.0,81.0)(12810.4,82.0)(12890.2,83.0)(12891.4,84.0)(12942.1,85.0)(13045.3,86.0)(13092.2,87.0)(13125.5,88.0)(13230.2,89.0)(13335.3,90.0)(13402.9,91.0)(13499.2,92.0)(13811.1,93.0)(14062.0,94.0)(14730.0,95.0)(14750.1,96.0)(14926.2,97.0)(15012.6,98.0)(15177.8,99.0)(15386.5,100.0)};
    \end{axis}
    \begin{axis}[
            name=doublebt10,
            title=(g) Double BT 10,
            at=(bugtrap10.below south west), anchor=above north west,
            xmin=0,
            xmax=46320.9,
        ]
        \addplot[color=blue]
        coordinates {(21700.4,1.0)(22026.5,2.0)(22581.3,3.0)(22595.3,4.0)(22845.5,5.0)(22860.0,6.0)(22960.2,7.000000000000001)(23288.0,8.0)(23291.6,9.0)(23367.9,10.0)(23527.4,11.0)(23541.8,12.0)(23593.1,13.0)(23639.2,14.000000000000002)(23661.5,15.0)(23710.6,16.0)(23737.1,17.0)(23752.0,18.0)(23799.1,19.0)(23811.0,20.0)(23958.8,21.0)(24018.7,22.0)(24355.9,23.0)(24389.0,24.0)(24425.5,25.0)(24462.2,26.0)(24518.2,27.0)(24598.6,28.000000000000004)(24714.4,28.999999999999996)(24742.5,30.0)(24798.5,31.0)(24809.8,32.0)(24866.1,33.0)(24905.3,34.0)(24980.9,35.0)(25181.9,36.0)(25268.0,37.0)(25309.2,38.0)(25326.6,39.0)(25379.2,40.0)(25392.4,41.0)(25393.3,42.0)(25443.1,43.0)(25482.1,44.0)(25515.0,45.0)(25544.8,46.0)(25575.7,47.0)(25576.2,48.0)(25597.5,49.0)(25607.0,50.0)(25714.8,51.0)(25828.1,52.0)(25841.1,53.0)(25868.8,54.0)(25889.0,55.00000000000001)(25994.0,56.00000000000001)(26049.8,56.99999999999999)(26122.7,57.99999999999999)(26123.7,59.0)(26159.6,60.0)(26175.1,61.0)(26204.2,62.0)(26216.7,63.0)(26246.3,64.0)(26249.8,65.0)(26312.6,66.0)(26325.3,67.0)(26342.9,68.0)(26379.9,69.0)(26401.3,70.0)(26422.1,71.0)(26526.0,72.0)(26558.3,73.0)(26616.1,74.0)(26673.9,75.0)(26727.3,76.0)(26918.7,77.0)(26939.5,78.0)(27010.4,79.0)(27029.9,80.0)(27184.1,81.0)(27207.9,82.0)(27260.2,83.0)(27337.8,84.0)(27404.8,85.0)(27416.6,86.0)(27623.7,87.0)(27661.0,88.0)(27719.5,89.0)(27811.8,90.0)(27856.1,91.0)(27876.9,92.0)(27917.7,93.0)(28078.5,94.0)(28491.0,95.0)(28985.4,96.0)(29018.5,97.0)(29237.2,98.0)(29319.4,99.0)(31042.5,100.0)};
        \addplot[color=orange]
        coordinates {(21068.9,1.0)(21804.6,2.0)(22016.5,3.0)(22400.6,4.0)(22544.7,5.0)(22558.4,6.0)(22640.7,7.000000000000001)(22960.1,8.0)(22963.0,9.0)(23274.3,10.0)(23301.1,11.0)(23418.0,12.0)(23592.5,13.0)(23620.8,14.000000000000002)(23677.4,15.0)(23711.8,16.0)(23721.3,17.0)(23798.1,18.0)(23853.4,19.0)(24093.2,20.0)(24220.1,21.0)(24224.2,22.0)(24298.5,23.0)(24356.7,24.0)(24359.0,25.0)(24387.6,26.0)(24388.2,27.0)(24431.4,28.000000000000004)(24467.7,28.999999999999996)(24620.5,30.0)(24652.0,31.0)(24805.0,32.0)(24806.8,33.0)(24874.8,34.0)(24947.2,35.0)(24961.4,36.0)(24987.0,37.0)(25085.8,38.0)(25170.4,39.0)(25241.4,40.0)(25269.7,41.0)(25281.6,42.0)(25306.9,43.0)(25340.7,44.0)(25393.8,45.0)(25515.5,46.0)(25530.0,47.0)(25530.7,48.0)(25540.7,49.0)(25654.2,50.0)(25684.2,51.0)(25712.8,52.0)(25890.4,53.0)(25939.4,54.0)(26174.8,55.00000000000001)(26180.7,56.00000000000001)(26268.6,56.99999999999999)(26680.4,57.99999999999999)(26735.8,59.0)(26827.0,60.0)(26858.7,61.0)(26907.0,62.0)(26960.3,63.0)(26965.3,64.0)(26972.7,65.0)(27165.1,66.0)(27168.7,67.0)(27187.2,68.0)(27188.8,69.0)(27216.7,70.0)(27289.7,71.0)(27318.1,72.0)(27494.6,73.0)(27564.5,74.0)(27582.8,75.0)(27612.4,76.0)(27660.2,77.0)(27810.8,78.0)(27884.7,79.0)(27968.3,80.0)(28050.4,81.0)(28069.5,82.0)(28073.4,83.0)(28145.3,84.0)(28576.4,85.0)(28644.9,86.0)(28688.5,87.0)(28836.7,88.0)(28965.9,89.0)(29109.3,90.0)(29132.4,91.0)(29218.1,92.0)(29275.4,93.0)(29729.0,94.0)(29861.9,95.0)(30254.1,96.0)(30385.7,97.0)(30403.5,98.0)(32212.8,99.0)(32576.1,100.0)};
        \addplot[color=green!60!black]
        coordinates {(30649.1,1.0)(30681.2,2.0)(31772.3,3.0)(32594.0,4.0)(32845.3,5.0)(33636.7,6.0)(33665.2,7.000000000000001)(33681.7,8.0)(33724.6,9.0)(34049.3,10.0)(34187.9,11.0)(34295.1,12.0)(34309.8,13.0)(34372.0,14.000000000000002)(34445.3,15.0)(34491.5,16.0)(34627.3,17.0)(34838.8,18.0)(34878.4,19.0)(34906.7,20.0)(34924.0,21.0)(34955.6,22.0)(35004.7,23.0)(35057.8,24.0)(35373.6,25.0)(35503.5,26.0)(35580.4,27.0)(35750.0,28.000000000000004)(35860.5,28.999999999999996)(35890.0,30.0)(36047.1,31.0)(36053.6,32.0)(36142.4,33.0)(36358.1,34.0)(36367.3,35.0)(36427.3,36.0)(36436.3,37.0)(36472.4,38.0)(36711.0,39.0)(36813.2,40.0)(36857.2,41.0)(36862.6,42.0)(37016.4,43.0)(37123.4,44.0)(37519.8,45.0)(37526.4,46.0)(37527.9,47.0)(37539.3,48.0)(37540.6,49.0)(37749.2,50.0)(37844.2,51.0)(38017.1,52.0)(38073.6,53.0)(38164.5,54.0)(38285.4,55.00000000000001)(38635.1,56.00000000000001)(38666.6,56.99999999999999)(38822.3,57.99999999999999)(38826.3,59.0)(39146.5,60.0)(39253.8,61.0)(39263.8,62.0)(39402.9,63.0)(39417.4,64.0)(39469.4,65.0)(39511.0,66.0)(39707.6,67.0)(39864.7,68.0)(39869.7,69.0)(39983.3,70.0)(40035.2,71.0)(40245.5,72.0)(40305.4,73.0)(40440.9,74.0)(40467.9,75.0)(40554.1,76.0)(40573.4,77.0)(40661.2,78.0)(40742.2,79.0)(40901.8,80.0)(40957.1,81.0)(40973.0,82.0)(41023.9,83.0)(41058.4,84.0)(41507.9,85.0)(41553.9,86.0)(41946.6,87.0)(42016.1,88.0)(42377.0,89.0)(42454.2,90.0)(42528.6,91.0)(42773.5,92.0)(42964.3,93.0)(43562.5,94.0)(44407.5,95.0)(45226.1,96.0)(45370.2,97.0)(45459.8,98.0)(45806.3,99.0)(46320.9,100.0)};
    \end{axis}
    \begin{axis}[
            name=potholes20,
            title=(b) Potholes 20,
            at=(potholes10.right of south east), anchor=left of south west,
            xmin=0,
            xmax=14560.5,
        ]
        \addplot[color=blue]
        coordinates {(4703.42,1.0)(5456.18,2.0)(5501.9,3.0)(5517.04,4.0)(5553.22,5.0)(5654.54,6.0)(5685.42,7.000000000000001)(5739.13,8.0)(5752.89,9.0)(5781.67,10.0)(5789.19,11.0)(5976.18,12.0)(6002.24,13.0)(6035.5,14.000000000000002)(6040.77,15.0)(6052.87,16.0)(6148.51,17.0)(6187.25,18.0)(6259.77,19.0)(6260.32,20.0)(6263.98,21.0)(6274.56,22.0)(6288.2,23.0)(6320.52,24.0)(6326.82,25.0)(6340.01,26.0)(6380.13,27.0)(6401.7,28.000000000000004)(6442.51,28.999999999999996)(6463.09,30.0)(6486.82,31.0)(6489.8,32.0)(6492.23,33.0)(6504.98,34.0)(6506.53,35.0)(6538.79,36.0)(6551.9,37.0)(6553.65,38.0)(6581.89,39.0)(6589.34,40.0)(6595.5,41.0)(6611.2,42.0)(6641.47,43.0)(6689.97,44.0)(6736.44,45.0)(6751.5,46.0)(6777.44,47.0)(6780.51,48.0)(6808.39,49.0)(6871.07,50.0)(6885.16,51.0)(6905.18,52.0)(6959.86,53.0)(7093.04,54.0)(7163.79,55.00000000000001)(7206.71,56.00000000000001)(7255.25,56.99999999999999)(7308.74,57.99999999999999)(7362.46,59.0)(7380.14,60.0)(7401.29,61.0)(7409.8,62.0)(7413.66,63.0)(7417.14,64.0)(7533.62,65.0)(7554.26,66.0)(7590.91,67.0)(7659.83,68.0)(7708.32,69.0)(7807.97,70.0)(7892.13,71.0)(7930.63,72.0)(7943.31,73.0)(8008.54,74.0)(8014.58,75.0)(8028.38,76.0)(8100.91,77.0)(8106.84,78.0)(8109.67,79.0)(8141.41,80.0)(8209.72,81.0)(8273.3,82.0)(8306.19,83.0)(8319.76,84.0)(8328.86,85.0)(8377.75,86.0)(8465.86,87.0)(8486.08,88.0)(8554.67,89.0)(8765.32,90.0)(8856.12,91.0)(9048.42,92.0)(9889.44,93.0)};
        \addplot[color=orange]
        coordinates {(6411.08,1.0)(6696.09,2.0)(6756.19,3.0)(6850.33,4.0)(7870.67,5.0)(7915.17,6.0)(7986.48,7.000000000000001)(8216.82,8.0)(8257.53,9.0)(8314.71,10.0)(8360.12,11.0)(8380.46,12.0)(8439.49,13.0)(8476.38,14.000000000000002)(8522.64,15.0)(8660.61,16.0)(8697.21,17.0)(8703.78,18.0)(8729.4,19.0)(8790.15,20.0)(8879.12,21.0)(8908.44,22.0)(9033.95,23.0)(9082.32,24.0)(9088.24,25.0)(9124.13,26.0)(9172.91,27.0)(9309.08,28.000000000000004)(9326.04,28.999999999999996)(9366.08,30.0)(9376.55,31.0)(9398.96,32.0)(9408.34,33.0)(9470.33,34.0)(9476.82,35.0)(9551.92,36.0)(9580.2,37.0)(9612.98,38.0)(9683.97,39.0)(9728.98,40.0)(9857.78,41.0)(9906.85,42.0)(9946.72,43.0)(9960.75,44.0)(9969.9,45.0)(10015.6,46.0)(10054.1,47.0)(10057.2,48.0)(10069.7,49.0)(10117.7,50.0)(10148.1,51.0)(10152.9,52.0)(10292.7,53.0)(10408.2,54.0)(10410.0,55.00000000000001)(10414.3,56.00000000000001)(10496.6,56.99999999999999)(10498.8,57.99999999999999)(10547.6,59.0)(10662.3,60.0)(10768.2,61.0)(10817.2,62.0)(10865.8,63.0)(10910.6,64.0)(10949.7,65.0)(10973.5,66.0)(10973.5,67.0)(11009.4,68.0)(11186.5,69.0)(11334.8,70.0)(11384.0,71.0)(11406.4,72.0)(11647.6,73.0)(11667.4,74.0)(11671.5,75.0)(11710.4,76.0)(11738.8,77.0)(11803.6,78.0)(11948.1,79.0)(12026.4,80.0)(12055.7,81.0)(12170.4,82.0)(12359.3,83.0)(12410.8,84.0)(12483.3,85.0)(12652.9,86.0)(12745.0,87.0)(12871.1,88.0)(13084.9,89.0)(13150.6,90.0)(13185.9,91.0)(13626.9,92.0)(13637.9,93.0)(13717.4,94.0)(13761.1,95.0)(13908.1,96.0)(14410.3,97.0)(14560.5,98.0)};
        \addplot[color=green!60!black]
        coordinates {(9057.92,1.0)(9384.76,2.0)(9385.45,3.0)(9417.38,4.0)(9472.19,5.0)(9503.38,6.0)(9515.26,7.000000000000001)(9531.1,8.0)(9638.64,9.0)(9711.3,10.0)(9818.85,11.0)(9841.93,12.0)(9921.57,13.0)(9929.38,14.000000000000002)(9981.96,15.0)(9991.47,16.0)(9997.52,17.0)(10067.2,18.0)(10079.1,19.0)(10092.8,20.0)(10112.6,21.0)(10145.5,22.0)(10193.0,23.0)(10196.4,24.0)(10201.6,25.0)(10273.0,26.0)(10329.1,27.0)(10379.8,28.000000000000004)(10431.6,28.999999999999996)(10446.7,30.0)(10466.0,31.0)(10485.8,32.0)(10493.0,33.0)(10557.8,34.0)(10568.8,35.0)(10630.3,36.0)(10645.0,37.0)(10650.7,38.0)(10675.1,39.0)(10682.0,40.0)(10683.4,41.0)(10696.8,42.0)(10731.6,43.0)(10794.4,44.0)(10807.3,45.0)(10812.1,46.0)(10814.0,47.0)(10831.4,48.0)(10864.5,49.0)(10944.8,50.0)(10954.7,51.0)(10967.5,52.0)(11046.2,53.0)(11049.7,54.0)(11059.7,55.00000000000001)(11060.1,56.00000000000001)(11084.3,56.99999999999999)(11101.3,57.99999999999999)(11116.1,59.0)(11119.1,60.0)(11186.2,61.0)(11214.8,62.0)(11229.0,63.0)(11312.7,64.0)(11327.1,65.0)(11386.1,66.0)(11398.2,67.0)(11424.7,68.0)(11434.2,69.0)(11508.9,70.0)(11511.7,71.0)(11512.9,72.0)(11559.5,73.0)(11622.4,74.0)(11721.5,75.0)(11765.2,76.0)(11775.5,77.0)(11867.3,78.0)(11886.1,79.0)(11943.2,80.0)(11951.9,81.0)(11956.0,82.0)(11974.6,83.0)(12002.9,84.0)(12014.0,85.0)(12114.5,86.0)(12120.2,87.0)(12136.7,88.0)(12157.7,89.0)(12207.3,90.0)(12259.0,91.0)(12326.8,92.0)(12353.5,93.0)(12445.4,94.0)(12454.7,95.0)(12468.2,96.0)(12553.8,97.0)(12757.6,98.0)(12766.9,99.0)(13323.3,100.0)};
    \end{axis}
    \begin{axis}[
            name=floor20,
            title=(d) Floor 20,
            at=(potholes20.below south west), anchor=above north west,
            xmin=0,
            xmax=39470.2,
        ]
        \addplot[color=blue]
        coordinates {(20941.4,1.0)(21299.1,2.0)(21327.6,3.0)(21693.2,4.0)(22207.6,5.0)(22328.9,6.0)(22407.5,7.000000000000001)(22505.8,8.0)(22560.8,9.0)(22669.0,10.0)(22695.8,11.0)(22804.2,12.0)(22898.9,13.0)(22904.4,14.000000000000002)(22905.1,15.0)(22921.0,16.0)(22950.1,17.0)(22956.4,18.0)(23049.6,19.0)(23053.9,20.0)(23087.4,21.0)(23088.7,22.0)(23221.4,23.0)(23267.0,24.0)(23270.0,25.0)(23331.1,26.0)(23356.4,27.0)(23380.2,28.000000000000004)(23421.5,28.999999999999996)(23433.9,30.0)(23531.2,31.0)(23542.0,32.0)(23545.9,33.0)(23579.5,34.0)(23590.5,35.0)(23609.3,36.0)(23630.1,37.0)(23647.1,38.0)(23708.0,39.0)(23708.6,40.0)(23753.7,41.0)(23760.8,42.0)(23765.6,43.0)(23847.1,44.0)(23891.6,45.0)(23945.0,46.0)(23949.2,47.0)(23952.3,48.0)(24037.7,49.0)(24056.9,50.0)(24076.4,51.0)(24077.4,52.0)(24077.4,53.0)(24168.3,54.0)(24219.2,55.00000000000001)(24225.7,56.00000000000001)(24229.2,56.99999999999999)(24237.8,57.99999999999999)(24321.0,59.0)(24375.6,60.0)(24377.9,61.0)(24427.3,62.0)(24567.8,63.0)(24594.3,64.0)(24602.7,65.0)(24657.2,66.0)(24667.4,67.0)(24695.3,68.0)(24702.7,69.0)(24751.3,70.0)(24759.2,71.0)(24763.6,72.0)(24842.4,73.0)(25035.6,74.0)(25087.8,75.0)(25089.5,76.0)(25195.8,77.0)(25235.3,78.0)(25344.1,79.0)(25376.9,80.0)(25597.2,81.0)(25603.5,82.0)(25697.4,83.0)(25782.5,84.0)(25845.9,85.0)(25846.4,86.0)(25890.9,87.0)(26012.8,88.0)(26027.9,89.0)(26164.8,90.0)(26646.8,91.0)(26704.8,92.0)(26872.8,93.0)(26980.3,94.0)(27058.4,95.0)(27097.4,96.0)(27254.4,97.0)(27614.8,98.0)(28333.2,99.0)(28797.7,100.0)};
        \addplot[color=orange]
        coordinates {};
        \addplot[color=green!60!black]
        coordinates {(24057.0,1.0)(24121.5,2.0)(24157.5,3.0)(24260.6,4.0)(24306.5,5.0)(24711.3,6.0)(25813.7,7.000000000000001)(25831.4,8.0)(25938.2,9.0)(26264.3,10.0)(26266.4,11.0)(26297.9,12.0)(26367.3,13.0)(26608.2,14.000000000000002)(26626.2,15.0)(26818.7,16.0)(26932.2,17.0)(26955.7,18.0)(27192.9,19.0)(27199.5,20.0)(27351.1,21.0)(27363.0,22.0)(27395.7,23.0)(27504.5,24.0)(27571.7,25.0)(27584.1,26.0)(28092.3,27.0)(28109.8,28.000000000000004)(28127.0,28.999999999999996)(28229.7,30.0)(28324.1,31.0)(28370.0,32.0)(28407.5,33.0)(28422.9,34.0)(28425.5,35.0)(28559.1,36.0)(28592.0,37.0)(28593.4,38.0)(28824.4,39.0)(28916.0,40.0)(29005.1,41.0)(29133.1,42.0)(29246.7,43.0)(29275.5,44.0)(29304.9,45.0)(29428.7,46.0)(29660.6,47.0)(29912.1,48.0)(29941.8,49.0)(30043.4,50.0)(30056.6,51.0)(30294.0,52.0)(30515.3,53.0)(30562.7,54.0)(30608.6,55.00000000000001)(30616.2,56.00000000000001)(30790.3,56.99999999999999)(31083.8,57.99999999999999)(31151.5,59.0)(31152.2,60.0)(31359.4,61.0)(31397.6,62.0)(31435.4,63.0)(31497.4,64.0)(31535.0,65.0)(31545.4,66.0)(31549.5,67.0)(31701.6,68.0)(31779.9,69.0)(31814.0,70.0)(31850.6,71.0)(31934.9,72.0)(31966.7,73.0)(32008.3,74.0)(32138.2,75.0)(32285.3,76.0)(32337.2,77.0)(32652.3,78.0)(32857.8,79.0)(33132.3,80.0)(33138.1,81.0)(33148.6,82.0)(33309.9,83.0)(33444.8,84.0)(33526.1,85.0)(33574.7,86.0)(33606.4,87.0)(33679.0,88.0)(33717.6,89.0)(33728.9,90.0)(33889.0,91.0)(33971.8,92.0)(34145.4,93.0)(34899.5,94.0)(35060.5,95.0)(35110.9,96.0)(35199.1,97.0)(36045.9,98.0)(37291.1,99.0)(39470.2,100.0)};
    \end{axis}
    \begin{axis}[
            name=bugtrap20,
            title=(f) Bugtrap 20,
            at=(floor20.below south west), anchor=above north west,
            xmin=0,
            xmax=18963.5,
        ]
        \addplot[color=blue]
        coordinates {(5355.15,1.0)(5581.63,2.0)(5751.07,3.0)(5776.01,4.0)(5781.53,5.0)(5852.02,6.0)(5863.29,7.000000000000001)(5929.15,8.0)(6102.77,9.0)(6122.02,10.0)(6125.21,11.0)(6153.69,12.0)(6161.46,13.0)(6168.49,14.000000000000002)(6193.82,15.0)(6206.18,16.0)(6240.53,17.0)(6252.92,18.0)(6270.35,19.0)(6273.52,20.0)(6281.39,21.0)(6308.31,22.0)(6347.3,23.0)(6370.06,24.0)(6393.03,25.0)(6426.04,26.0)(6474.26,27.0)(6486.56,28.000000000000004)(6502.63,28.999999999999996)(6539.21,30.0)(6568.52,31.0)(6580.89,32.0)(6613.19,33.0)(6669.44,34.0)(6691.78,35.0)(6697.56,36.0)(6714.07,37.0)(6742.32,38.0)(6820.55,39.0)(6848.86,40.0)(6908.75,41.0)(6927.32,42.0)(6928.05,43.0)(6971.93,44.0)(6984.23,45.0)(6993.68,46.0)(6997.98,47.0)(6997.98,48.0)(7004.18,49.0)(7014.01,50.0)(7025.46,51.0)(7028.79,52.0)(7033.35,53.0)(7056.17,54.0)(7088.33,55.00000000000001)(7120.12,56.00000000000001)(7146.95,56.99999999999999)(7219.74,57.99999999999999)(7222.66,59.0)(7245.15,60.0)(7253.98,61.0)(7261.17,62.0)(7276.26,63.0)(7320.35,64.0)(7343.45,65.0)(7358.31,66.0)(7420.52,67.0)(7439.68,68.0)(7456.66,69.0)(7488.68,70.0)(7516.07,71.0)(7556.05,72.0)(7563.08,73.0)(7565.63,74.0)(7582.72,75.0)(7634.2,76.0)(7682.31,77.0)(7691.89,78.0)(7695.61,79.0)(7708.91,80.0)(7773.98,81.0)(7781.99,82.0)(7792.76,83.0)(7797.94,84.0)(7799.0,85.0)(7803.58,86.0)(7811.85,87.0)(7815.65,88.0)(7865.51,89.0)(7898.45,90.0)(7918.75,91.0)(7985.37,92.0)(8126.2,93.0)(8131.74,94.0)(8509.47,95.0)(8819.71,96.0)(8976.77,97.0)(9176.8,98.0)(9225.69,99.0)(9952.23,100.0)};
        \addplot[color=orange]
        coordinates {(7200.21,1.0)(7281.03,2.0)(7309.66,3.0)(7373.26,4.0)(7454.63,5.0)(7683.31,6.0)(7728.55,7.000000000000001)(7742.91,8.0)(7986.73,9.0)(7992.56,10.0)(8021.32,11.0)(8190.9,12.0)(8250.75,13.0)(8260.06,14.000000000000002)(8360.95,15.0)(8387.57,16.0)(8631.37,17.0)(8701.78,18.0)(8749.65,19.0)(8837.46,20.0)(8842.31,21.0)(8979.33,22.0)(8986.95,23.0)(9021.5,24.0)(9029.54,25.0)(9058.86,26.0)(9115.81,27.0)(9137.71,28.000000000000004)(9163.24,28.999999999999996)(9347.74,30.0)(9365.41,31.0)(9378.45,32.0)(9392.34,33.0)(9412.48,34.0)(9413.59,35.0)(9444.6,36.0)(9488.13,37.0)(9571.32,38.0)(9680.92,39.0)(9718.35,40.0)(9775.86,41.0)(9868.95,42.0)(9875.3,43.0)(9904.34,44.0)(9911.7,45.0)(9948.49,46.0)(9999.85,47.0)(10002.1,48.0)(10012.9,49.0)(10046.7,50.0)(10143.4,51.0)(10160.5,52.0)(10288.2,53.0)(10296.7,54.0)(10324.1,55.00000000000001)(10358.1,56.00000000000001)(10447.4,56.99999999999999)(10468.0,57.99999999999999)(10478.9,59.0)(10672.0,60.0)(10701.1,61.0)(10774.7,62.0)(10776.9,63.0)(10814.1,64.0)(10823.4,65.0)(10827.2,66.0)(10970.8,67.0)(11060.5,68.0)(11082.7,69.0)(11159.8,70.0)(11478.4,71.0)(11501.7,72.0)(11641.5,73.0)(11704.5,74.0)(11774.1,75.0)(12050.0,76.0)(12148.6,77.0)(12317.4,78.0)(12717.0,79.0)(12751.2,80.0)(12980.1,81.0)(13050.7,82.0)(13132.9,83.0)(13151.0,84.0)(13265.6,85.0)(13315.3,86.0)(13381.1,87.0)(13711.4,88.0)(13776.2,89.0)(13778.0,90.0)(13894.0,91.0)(14307.8,92.0)(14629.6,93.0)(14894.2,94.0)(15021.7,95.0)(15198.2,96.0)(15199.5,97.0)(15555.5,98.0)(16603.1,99.0)(18963.5,100.0)};
        \addplot[color=green!60!black]
        coordinates {(11200.0,1.0)(11708.5,2.0)(12145.6,3.0)(12210.9,4.0)(12231.1,5.0)(12241.3,6.0)(12247.6,7.000000000000001)(12296.5,8.0)(12472.3,9.0)(12533.8,10.0)(12546.4,11.0)(12584.6,12.0)(12588.4,13.0)(12598.1,14.000000000000002)(12722.8,15.0)(12773.3,16.0)(12827.0,17.0)(12837.6,18.0)(12852.6,19.0)(12897.4,20.0)(12906.1,21.0)(12948.2,22.0)(12999.2,23.0)(13032.3,24.0)(13037.1,25.0)(13067.7,26.0)(13081.0,27.0)(13082.9,28.000000000000004)(13105.1,28.999999999999996)(13241.1,30.0)(13267.3,31.0)(13284.7,32.0)(13296.8,33.0)(13310.7,34.0)(13337.0,35.0)(13355.2,36.0)(13409.8,37.0)(13443.7,38.0)(13451.5,39.0)(13495.7,40.0)(13516.2,41.0)(13519.1,42.0)(13540.0,43.0)(13571.0,44.0)(13583.4,45.0)(13623.5,46.0)(13677.2,47.0)(13685.1,48.0)(13705.8,49.0)(13753.5,50.0)(13836.4,51.0)(13843.3,52.0)(13875.3,53.0)(14030.2,54.0)(14039.9,55.00000000000001)(14042.9,56.00000000000001)(14046.2,56.99999999999999)(14100.3,57.99999999999999)(14108.6,59.0)(14144.5,60.0)(14147.7,61.0)(14168.6,62.0)(14196.3,63.0)(14230.1,64.0)(14284.7,65.0)(14300.6,66.0)(14316.0,67.0)(14349.7,68.0)(14384.2,69.0)(14399.9,70.0)(14426.6,71.0)(14495.5,72.0)(14500.1,73.0)(14512.3,74.0)(14514.1,75.0)(14524.3,76.0)(14606.5,77.0)(14664.9,78.0)(14685.1,79.0)(14749.3,80.0)(14778.4,81.0)(14787.3,82.0)(14888.4,83.0)(14914.3,84.0)(14982.1,85.0)(15064.7,86.0)(15069.7,87.0)(15180.2,88.0)(15204.4,89.0)(15277.7,90.0)(15375.6,91.0)(15376.0,92.0)(15495.0,93.0)(15519.3,94.0)(15563.0,95.0)(15905.5,96.0)(15991.4,97.0)(16434.6,98.0)(16568.4,99.0)(17067.7,100.0)};
    \end{axis}
    \begin{axis}[
            name=doublebt20,
            title=(h) Double BT 20,
            at=(bugtrap20.below south west), anchor=above north west,
            xmin=0,
            xmax=49780.2
        ]
        \addplot[color=blue]
        coordinates {(22854.9,1.0)(23496.4,2.0)(24523.9,3.0)(24578.0,4.0)(24936.0,5.0)(24945.6,6.0)(24986.6,7.000000000000001)(25106.1,8.0)(25212.5,9.0)(25225.2,10.0)(25386.9,11.0)(25534.4,12.0)(25596.0,13.0)(25663.4,14.000000000000002)(25700.8,15.0)(26075.7,16.0)(26154.3,17.0)(26180.8,18.0)(26206.6,19.0)(26226.3,20.0)(26327.5,21.0)(26341.2,22.0)(26358.9,23.0)(26471.6,24.0)(26524.3,25.0)(26529.2,26.0)(26542.5,27.0)(26635.4,28.000000000000004)(26658.9,28.999999999999996)(26676.7,30.0)(26687.6,31.0)(26709.2,32.0)(26741.4,33.0)(26817.7,34.0)(26844.0,35.0)(26861.6,36.0)(26868.7,37.0)(26926.4,38.0)(26946.9,39.0)(26977.5,40.0)(26987.7,41.0)(27064.2,42.0)(27178.0,43.0)(27182.3,44.0)(27188.5,45.0)(27246.2,46.0)(27293.1,47.0)(27306.6,48.0)(27328.4,49.0)(27464.7,50.0)(27468.5,51.0)(27480.3,52.0)(27604.3,53.0)(27649.1,54.0)(27691.7,55.00000000000001)(27736.5,56.00000000000001)(27740.0,56.99999999999999)(27753.9,57.99999999999999)(27821.1,59.0)(27943.8,60.0)(28259.3,61.0)(28347.4,62.0)(28371.9,63.0)(28438.1,64.0)(28494.9,65.0)(28574.0,66.0)(28663.5,67.0)(28762.6,68.0)(28837.3,69.0)(28896.4,70.0)(29003.7,71.0)(29009.0,72.0)(29116.1,73.0)(29356.5,74.0)(29477.4,75.0)(29629.9,76.0)(29639.9,77.0)(29642.3,78.0)(29648.2,79.0)(29853.2,80.0)(30050.5,81.0)(30089.4,82.0)(30194.8,83.0)(30206.1,84.0)(30209.7,85.0)(30311.2,86.0)(30404.7,87.0)(30554.0,88.0)(30883.1,89.0)(31171.9,90.0)(31554.6,91.0)(31671.0,92.0)(31702.8,93.0)(32008.1,94.0)(32406.7,95.0)(32620.9,96.0)(32923.1,97.0)(33170.8,98.0)(33542.4,99.0)(34897.4,100.0)};
        \addplot[color=orange]
        coordinates {(31505.8,1.0)(32866.1,2.0)(35643.3,3.0)};
        \addplot[color=green!60!black]
        coordinates {(31260.1,1.0)(31489.9,2.0)(33591.8,3.0)(33937.4,4.0)(34034.2,5.0)(34076.7,6.0)(34131.4,7.000000000000001)(34197.4,8.0)(34546.3,9.0)(34586.5,10.0)(34786.6,11.0)(35080.5,12.0)(35157.8,13.0)(35177.3,14.000000000000002)(35179.0,15.0)(35225.5,16.0)(35562.3,17.0)(35642.4,18.0)(35681.6,19.0)(35740.2,20.0)(35913.8,21.0)(36029.8,22.0)(36307.3,23.0)(36656.2,24.0)(36711.7,25.0)(36791.3,26.0)(36901.1,27.0)(36906.7,28.000000000000004)(36926.1,28.999999999999996)(37078.0,30.0)(37129.1,31.0)(37239.0,32.0)(37789.6,33.0)(37790.8,34.0)(37845.9,35.0)(37857.9,36.0)(37965.2,37.0)(37979.1,38.0)(38039.0,39.0)(38143.2,40.0)(38160.7,41.0)(38218.7,42.0)(38369.1,43.0)(38418.9,44.0)(38620.1,45.0)(38651.8,46.0)(38710.5,47.0)(38752.7,48.0)(38765.1,49.0)(38808.2,50.0)(38909.8,51.0)(39018.2,52.0)(39124.5,53.0)(39207.0,54.0)(39560.6,55.00000000000001)(39852.1,56.00000000000001)(40018.7,56.99999999999999)(40119.6,57.99999999999999)(40120.4,59.0)(40257.7,60.0)(40281.5,61.0)(40325.3,62.0)(40433.0,63.0)(40445.9,64.0)(40490.7,65.0)(40714.6,66.0)(40739.5,67.0)(40888.6,68.0)(40957.0,69.0)(40966.5,70.0)(40979.9,71.0)(41010.2,72.0)(41014.1,73.0)(41077.2,74.0)(41146.8,75.0)(41160.0,76.0)(41194.3,77.0)(41202.8,78.0)(41474.2,79.0)(41653.8,80.0)(41825.5,81.0)(42076.6,82.0)(42300.1,83.0)(42338.1,84.0)(42479.1,85.0)(42606.7,86.0)(43390.2,87.0)(43403.3,88.0)(43416.1,89.0)(43457.8,90.0)(43707.3,91.0)(44449.5,92.0)(44707.7,93.0)(44907.6,94.0)(45512.5,95.0)(46657.5,96.0)(46865.9,97.0)(47833.8,98.0)(49049.6,99.0)(49780.2,100.0)};
    \end{axis}
\end{tikzpicture}